\documentclass[12pt]{elsarticle}

\usepackage{amsmath, amsfonts, amssymb}
\usepackage[margin=0.75in]{geometry}
\usepackage{graphicx}
\usepackage{hyperref}
\usepackage{setspace}
\usepackage{algorithm}
\usepackage{algpseudocode}
\usepackage{tabularx}
\usepackage{lscape} %vertical table
\usepackage{rotating}
\usepackage{fancyhdr}
\usepackage{hyphenat} 
\usepackage{enumitem}
\usepackage{mathrsfs}
\usepackage{bm}
\usepackage{placeins} %To restrict images inside sections/subsections
\usepackage{cleveref}
\usepackage{fancyhdr}
\usepackage{color}
\usepackage{lineno}
\usepackage{subcaption}
\usepackage{multirow}
\usepackage{adjustbox} %To change table preferences
\usepackage{tikz} %To add checkmark
\usepackage{longtable} %Tables spanning multiple pages
\usepackage{caption}
\usepackage{graphicx}
\usepackage{makecell}
\usepackage{natbib}
\setcitestyle{square, comma, numbers, compress, super}
\usepackage{array, booktabs, makecell} %For breaking table header into multiple lines

\journal{XXXXXX}

\begin{document}

\begin{frontmatter}

\title{Differentiable Autoencoding Neural Operator for Interpretable and Integrable Latent Space Modeling}


\author[utah1,utah2]{Siva Viknesh}
\author[utah1,utah2]{Amirhossein Arzani\corref{cor1}} %% Author name
\ead{amir.arzani@sci.utah.edu}

\address[utah1]{Department of Mechanical Engineering, University of Utah, Salt Lake City, UT, USA}
\address[utah2]{Scientific Computing and Imaging Institute, University of Utah, Salt Lake City, UT, USA}
\cortext[cor1]{Corresponding author}

\begin{abstract}
Scientific machine learning has enabled the extraction of physical insights and data-driven modeling of high-dimensional spatiotemporal data, yet achieving physically interpretable latent representations and computationally efficient surrogates remains an open challenge. To address this, we propose the \textbf{DI}fferentiable \textbf{A}utoencoding \textbf{N}eural \textbf{O}perator - \textbf{DIANO}, a deterministic autoencoding neural operator framework that constructs visualizable coarse-grid latent spaces for both dimensional and geometric reduction across varying spatial discretizations, with governing equations enforced directly within the latent space. Built upon neural operators, DIANO achieves this through an encoding neural operator that spatially coarsens the high-dimensional input functions into the latent representation, and a decoding neural operator that reconstructs the original inputs via spatial refinement. We assess DIANO's latent representation and dimensionality reduction performance against baselines, including the Convolutional Neural Operator and standard autoencoders. Furthermore, a fully differentiable partial differential equation (PDE) solver is integrated as the sole input-output functional mapping operator within the latent space, enabling end-to-end training with governing physics prescribed a priori through parametric PDEs. Various PDE formulations are investigated, including the 2D unsteady advection-diffusion and the 3D Pressure--Poisson equation, revealing that the fidelity of the embedded PDE relative to the true governing physics is a critical design factor that jointly governs the learned latent representation and reconstruction accuracy. Benchmark problems include flow past a 2D cylinder, flow through a 2D symmetric stenosed artery, and a 3D patient-specific coronary artery, demonstrating accurate reconstruction of high-fidelity spatio-temporal fields through low-fidelity latent PDE evolution at reduced computational cost, while yielding coherent, spatially organized, and  meaningful latent structures.

\end{abstract}

\begin{keyword}
Interpretable machine learning \sep Dimensionality reduction \sep Differentiable PDE solver \sep Neural operators \sep Autoencoder
\end{keyword}



\end{frontmatter}

%\linenumbers

\section{Introduction}
\label{sec1}

Modeling and simulating complex physical systems such as fluid flows governed by nonlinear partial differential equations (PDEs) remains a fundamental challenge in computational physics and engineering. High-fidelity simulations of such flows generate extremely high-dimensional spatiotemporal datasets, leading to substantial demands on storage, computation, and limiting their applicability in real-time analysis, optimization, and control. To mitigate these constraints, reduced-order modeling (ROM) techniques aim to construct low-dimensional surrogate models that capture the essential dynamics while significantly reducing computational costs.

Classical linear ROM techniques, such as Proper Orthogonal Decomposition (POD)~\cite{lumley1967structure,berkooz1993proper} and Dynamic Mode Decomposition (DMD)~\cite{schmid2010dynamic}, achieve dimensionality reduction by projecting high-dimensional flow fields onto linear subspaces spanned by data-driven modes. While effective for moderately nonlinear systems, these methods often struggle with strongly nonlinear, multiscale behaviors typical of complex flows~\cite{zhang2015machine, ling2016reynolds, kutz2017deep, eivazi2020deep}.  This limitation has motivated a shift toward nonlinear dimensionality reduction and manifold learning techniques, which aim to uncover intrinsic low-dimensional latent manifolds that more faithfully represent the underlying physics. Such a manifold is a topological space that is locally diffeomorphic to a Euclidean space of a given dimension, which allows the complete system to be described using a reduced set of coordinates, while the ROM governs the dynamics evolving within this space. Manifold learning techniques such as Isometric Mapping (ISOMAP), t-distributed Stochastic Neighbor Embedding (t-SNE), Uniform Manifold Approximation and Projection (UMAP), kernel Principal Component Analysis (kPCA), and specifically deep autoencoder networks such as convolutional autoencoders (CNN-AE) and fully connected autoencoders (NN-AE) have demonstrated success in extracting compact nonlinear representations of complex flows~\cite{roweis2000nonlinear, eivazi2020deep, csala2022comparing, li2024manifold}. Among these, deep autoencoders have shown particular promise due to their capacity to learn task-specific embeddings via end-to-end training~\cite{erichson2020shallow, agostini2020exploration, fukami2020convolutional, fukami2021model, nakamura2021convolutional}. Variational autoencoders (VAEs) further extend this paradigm by incorporating probabilistic modeling, enabling generative modeling and uncertainty quantification~\cite{sakurada2014anomaly, cheng2020advanced, qu2021deep}.

Neural operator frameworks have emerged as powerful tools for learning mappings between infinite-dimensional function spaces. Representative architectures differ in their choice of functional basis, e.g., the Laplacian Neural Operator (LNO) leverages the Laplacian eigenfunctions~\cite{cao2024laplace}, the Fourier Neural Operator (FNO) utilizes global Fourier modes~\cite{li2020fourier}, the Convolutional Neural Operator (CNO) employs spatial convolutional kernels~\cite{raonic2023convolutional}, and the Deep Operator Network (DeepONet)~\cite{lu2019deeponet, lu2022comprehensive} adopts a branch–trunk neural architecture. Unlike classical neural networks defined on fixed discretizations, these neural operators approximate solution operators of PDEs in a mesh-independent manner, thereby enabling generalization across varying spatio-temporal resolutions and zero-shot super-resolution. Extensions to probabilistic settings include Generative Adversarial Networks (GAN)~\cite{rahman2022generative}, diffusion models~\cite{lim2023score}, and Variational Autoencoding Neural Operator (VANO)~\cite{seidman2023variational}. Notably, VANO is the first \textit{autoencoding} architecture in the realm of operator learning that simultaneously performs nonlinear dimensionality reduction (spatial coarsening and refinements) and functional mapping via a compact latent-space representation. Worth noting that there has been increased attention on developing geometry-aware neural operators, ~\cite{li2020multipole, li2022transformer, serrano2023operator, hao2023gnot, alkin2024universal, li2023geometry, han2025geomano}, which incorporate geometric or grid-dependent information to improve modeling performance on complex domains. Moreover, this family of neural operators, together with classical deep autoencoders, offers a methodologically integrated framework for projecting high-dimensional input data onto a latent manifold that encodes the intrinsic characteristics and dominant signatures of the governing dynamics. However, because the latent space is inherently distinct from the physical space, establishing a direct correspondence between latent representations and the underlying physical structures remains an open research problem. This limits interpretability and motivates the development of autoencoding methodologies that preserve a physically meaningful mapping between latent and physical spaces.

%Our recent work~\cite{csala2025physics} aligns with this direction by improving 1D blood flow models that typically approximate the cross-sectionally averaged dynamics of the full 3D blood flow equations. A physics-constrained Neural PDE framework has been developed that enables learning mappings across geometric dimensions (1D~$\leftarrow$~3D) by leveraging the underlying PDE structure rather than ML methods, although the latter are recommended for more complex geometries.

Another key challenge in ROMs lies in capturing the \textit{temporal evolution} of transient systems, which is crucial for accurate prediction of unsteady and nonlinear dynamics. In deep autoencoder-based models, latent-space temporal evolution is typically captured using recurrent neural networks such as Long Short-Term Memory (LSTM) networks~\cite{nakamura2021convolutional, bukka2021assessment, gupta2022three}, Neural Ordinary Differential Equation (NODE)~\cite{chen2018neural}, DMD~\cite{eivazi2020deep, zhang2023nonlinear}, and attention-based mechanisms~\cite{peng2022attention}. In the neural operator domain, several models have been proposed to incorporate spatio-temporal learning. Developments include the Wavelet Neural Operator (WNO)~\cite{tripura2022wavelet}, U-shaped Neural Operator~\cite{rahman2022u}, Neural Operators on Riemannian Manifolds (NORM)~\cite{chen2024learning}, and the Spatio-Temporal Neural Operator~\cite{fu2025spatio}. Additionally, attention-based~\cite{alkin2024universal, peng2023linear} and Recurrent Neural Operator models~\cite{ye2025recurrent} have been introduced to improve temporal modeling capabilities. Despite these advances, accurately and efficiently modeling nonlinear, multiscale, and time-evolving fluid dynamics remains an open research challenge. One promising approach lies in embedding \textit{physics priors}, such as conservation laws and governing differential equations, directly into machine learning (ML) architectures to produce physically consistent spatio-temporal predictions~\cite{karniadakis2021physics}. Broadly, these approaches fall into two categories: \textit{physics-constrained} and \textit{physics-invoked} models~\cite{faroughi2022physics}. Physics-constrained models~\cite{raissi2019physics, mattey2021physics, arzani2023theory, mohan2023embedding, chalapathi2024scaling, karnakov2024solving} incorporate physical laws \textit{explicitly} during training, typically through hard constraints or penalty terms that enforce PDE residuals, boundary conditions, or conservation properties, thereby guiding the model toward physically valid output solutions, improving predictive accuracy, and reducing data requirements. In contrast, physics-invoked models~\cite{wiewel2019latent, wu2022learning, brandstetter2022message, lippe2023pde, liu2024multi, li2025latent, fukami2023grasping, mousavi2025low, fukami2025observable} incorporate physical insight more \textit{implicitly}, leveraging architectural designs inspired by physical principles, including symmetry and locality, and leverage known flow quantities and probabilistic structures, without directly enforcing governing equations.

Beyond incorporating physical priors, the integration of ML optimization techniques into classical numerical PDE solvers has recently gained considerable attention. These approaches exploit the computational graph of the ML environment and help the PDE solvers to solve inverse problems and improve closure models in a problem-specific, data-driven manner, resulting in substantial enhancements in the accuracy and reliability of spatio-temporal solutions. The augmentation of classical numerical PDE solvers with ML architectures typically follows two main strategies: \textit{fully differentiable PDE solvers}~\cite{karnakov2024solving, boral2023neural, sirignano2023dynamic} and \textit{hybrid neural–physics solvers}~\cite{liu2024multi, kochkov2021machine, fan2024differentiable}. The primary distinction between these approaches lies in the nature of coupling. Hybrid methods generally employ \textit{weak coupling}, wherein ML models are trained offline and subsequently integrated into conventional solvers~\cite{tompson2017accelerating, duraisamy2019turbulence, vinuesa2022enhancing, margenberg2024dnn}. In contrast, fully differentiable PDE solvers adopt \textit{strong coupling}, simultaneously training ML components and solving PDEs via end-to-end computation. The latter method has seen increasing advocacy due to its ability to improve computational fidelity and adaptability~\cite{belbute2020combining, list2022learned, list2025differentiability, akhare2025implicit}. In both approaches, the solvers operate on either fully resolved or filtered (coarse-grained) governing PDEs, learning unknown physical parameters or closure models concurrently with the flow solution~\cite{wang2024beyond}. Complementary to these developments, \textit{physics discovery} in latent spaces has emerged through the Latent Space Dynamics Identification (LaSDI) framework~\cite{fries2022lasdi}. LaSDI constructs non-intrusive reduced-order models by encoding high-dimensional PDE solutions into a low-dimensional latent space, where temporal evolution is governed by learned ODEs. This approach yields compact and interpretable representations of system dynamics~\cite{li2022transformer, champion2019data, reinbold2019data, maulik2020time, huang2020learning, negiar2022learning}, though it may not strictly enforce physical laws. Recent extensions, such as Thermodynamics-LaSDI (T-LaSDI)~\cite{park2024tlasdi}, incorporate physics-based constraints to improve consistency and predictive accuracy. Overall, LaSDI provides a promising pathway for interpretable and efficient reduced-order modeling~\cite{bonneville2024comprehensive}.

Despite recent advances, no significant studies have explored embedding physical priors \textit{directly} into the latent space of ML models. Such integration is crucial for ensuring  that the latent dynamics evolve in strict accordance with the embedded governing equations, thereby mitigating long-term drift and enabling the spatio-temporal visualization of \textit{interpretable} latent representation. In the present work, ``interpretable latent representation'' is defined as a coarse-grid visualization of the latent space that may display coherent, spatially organized structures, as demonstrated later, depending on the choice of the physics imposed in the latent space. While physics-informed ML has made substantial progress, most existing approaches enforce physical constraints either at the level of the loss function or on reconstructed outputs, leaving latent spaces largely unconstrained and disconnected from the underlying dynamics. The primary challenge lies in the requirement for fully differentiable numerical solvers capable of propagating physical priors through all hidden representations. Recent advances in differentiable PDE solvers, however, have rendered this feasible. Notably, decoder-only architectures~\cite{dashtbayaz2025physics} have incorporated a differentiable solver at the decoder output, forcing that system dynamics can be faithfully mirrored within latent representations via decoder-invert operations. This approach enables latent spaces that both reflect the underlying physics and improve limited interpretability. A related approach employs a deep neural network–based autoencoder (NN-AE) for flow super-resolution~\cite{paliard2022exploring}. The encoder maps low-resolution data into a latent space, and the decoder reconstructs high-resolution fields. In this framework, the latent space first reshapes the encoder output into a grid-structured representation, which is advanced in time by an LSTM and subsequently \textit{corrected} with a differentiable solver. While the solver is embedded within the latent space, the temporal dynamics are still primarily learned by the LSTM and only improved by the solver.

%A key enabler of this design is the geometrical structure of the latent representation considered here. By construction, the latent space retains only the dominant, dynamically relevant features of the full-resolution field while filtering the fine-scale features, an informational reduction that simultaneously promotes visualization and computational tractability.

The present work departs from the prevailing paradigm of discovering models in the latent space from data. Rather, we deliberately prescribe the known governing physics, \textit{a priori}, directly within the latent space, thereby forcing the latent dynamics to conform to an imposed physical prior. This positions the proposed framework as a data-driven differentiable PDE solver, with end-to-end differentiability serving as the mechanism by which the physical prior is propagated throughout the architecture during training. 
This architectural design raises a fundamental feasibility question: given high-resolution data governed by a high-fidelity PDE, can the same class of problems be accurately solved by coarse-graining the input into a structured latent space, evolving it with a simplified low-fidelity form of the same governing PDE (for instance, through a linearized form), and recovering the high-resolution fields through decoding, entirely in an end-to-end differentiable manner? And if the answer is affirmative, how does the choice of latent PDE fidelity, relative to the underlying flow physics, govern the correspondence between coarse-grained latent structures and given high-resolution dynamics, and what are the quantitative consequences for reconstruction accuracy? Answering these questions demands a systematic investigation of the interplay between latent solver fidelity, the coarse-grained latent structure it induces, and the resulting reconstruction quality, a relationship that has not yet been rigorously examined in the literature. This coarse-grid structure makes the latent space a natural substrate for a differentiable PDE solver, enabling the governing equations to be advanced directly within the latent representation rather than over the full-resolution domain (and potentially with a PDE that is easier to solve). Unlike conventional PDE solvers, which require detailed resolution across all scales (from global structures to fine-grained fluctuations), making them computationally intensive, grid-based latent dynamics bypass this requirement, operating only over the semantically meaningful coarse-grained scales. Another practically significant consequence is that the \textit{fidelity} of the latent-embedded PDE solvers becomes a free design choice here, allowing the selection of any admissible simplified form of the governing high-fidelity equations and systematically balancing computational efficiency against modeling accuracy, a flexibility unavailable in conventional PDE solvers.

%The framework proposed herein constitutes a physics-embedded surrogate modeling strategy that enables computationally efficient and accurate approximation of high-fidelity PDE-governed solutions through deliberate, low-fidelity latent-space evolution. Beyond computational efficiency, it provides a principled basis for understanding how the fidelity of the embedded governing equations should be selected to achieve superior reconstruction quality -- a question of fundamental relevance to the broader effort of unifying data-driven learning with physics-based modeling.

%Given the growing demand for accurate and efficient modeling of high-dimensional, transient, and nonlinear flow fields, there is increasing interest in the development of ML methodologies that effectively integrate (a) mesh-invariant nonlinear dimensionality and geometrical reductions, (b) interpretable latent-space dynamics, and (c) physics-based latent temporal marching. Such frameworks aim to retain only the essential physical structures, enable their coherent evolution within a reduced latent space, and ensure that these representations remain both interpretable and consistent with the governing flow dynamics. 

In this work, we propose the \textbf{DI}fferentiable \textbf{A}utoencoding \textbf{N}eural \textbf{O}perator - \textbf{DIANO}, a deterministic framework that introduces physics-based spatio-temporal modeling on a coarse latent grid. DIANO seamlessly addresses the curse of dimensionality by employing nonlinear autoencoders to compress high-dimensional flow fields into low-dimensional latent representations defined on a coarsened grid, retaining only the essential and physically meaningful structures. Mesh-resolution invariance is achieved through neural operator architectures that learn mappings between input and output functions, thus enabling generalization across a wide range of spatial discretizations. A key innovation of DIANO lies in the integration of fully differentiable PDE solvers within the latent space, allowing the temporal evolution of latent variables to be governed directly by the underlying physical laws rather than purely data-driven ML approximations. Furthermore, generalization across PDE parameters is incorporated in a physically meaningful way by varying the parameters of the embedded latent PDE model directly, rather than augmenting them as additional external conditioning inputs. By tightly coupling dimensionality/geometrical reduction, operator learning, and PDE-based latent dynamics evolution, DIANO may offer a unified, scalable, and physics-embedded framework for modeling complex spatio-temporal flow phenomena. Specifically, our key contributions include:

\begin{itemize}
    \item \textbf{Deterministic Autoencoding Neural Operator:} Introduced a mesh-invariant operator framework that maps high-dimensional fields defined on a $N \times N$ grid to latent representations defined on a coarser grid ($M \times M$, $N > M$) and reconstructs them.
    \item \textbf{Visualizable Coarse-Grid Latent Representation:} Established a novel coarse-grid latent representation,  defined as a coarsened input grid, which enables visualization  of coarse-grained structures in the latent space. The latent structures may be physically interpretable based on the choice of the imposed latent PDE but will always be visualizable (even if not physically meaningful). 
    \item \textbf{Latent Space PDE Integration:} Embedded differentiable PDE solvers within the coarse-grid latent space, enabling efficient computations while maintaining physics-constrained temporal evolution. 
    \item \textbf{Flexible Solver-Accuracy Trade-offs:} \ Demonstrated that solving lower-fidelity PDEs, a simplified form of governing high-fidelity PDEs, in the latent space allows a flexible trade-off between differentiable solver complexity/efficiency and reconstruction accuracy.
    \item \textbf{A Paradigm Shift in Scientific Autoencoders:} The main goal of autoencoders in scientific machine learning is to discover a low-dimensional representation that facilitates downstream tasks and latent model discovery. We shift this paradigm by regularizing the latent space with a low-fidelity PDE (hybrid data-driven and physics-driven latent discovery), which has the benefit of developing efficient low-fidelity solvers and corrections in an end-to-end manner.   
    \item \textbf{Geometrical Reduction with Operator Learning:} Enabled operator learning to learn geometrical reductions, mapping of high-dimensional geometric data to lower-dimensional latent geometric data (e.g., 2D $\rightarrow$ 1D), while ensuring that the corresponding (1D) lower-dimensional PDE is solved on the reduced data, and then mapped back to the original high-dimensional geometric field.
\end{itemize}

The paper is organized as follows. Section~\ref{sec3} presents the problem formulation, outlining four modeling scenarios. It then introduces the proposed DIANO framework, describing the architectural variants for each scenario and the development of a differentiable PDE solver, followed by a discussion of the benchmark flow problems considered in this study. Section~\ref{sec4} analyzes the results of the DIANO framework, focusing on the spatio-temporal latent representations and their implications for quantitative mapping accuracy across the four scenarios. Section~\ref{sec5} provides a discussion of the results and highlights potential directions for future work. Finally, Section~\ref{sec6} concludes the study, summarizing the contributions and key findings of the DIANO framework.

\section{Methods}\label{sec3}
This section begins by presenting four representative modeling scenarios on which the proposed DIANO framework demonstrates its capabilities. We then introduce the DIANO framework, which employs Fourier layers within the autoencoding process to capture spatial representations. Finally, we describe the construction of a differentiable PDE solver that integrates the governing equations directly into the DIANO latent space, enabling latent temporal evolution.

\subsection{Problem Description}
\label{sec2}
We consider four representative problem configurations to demonstrate the capabilities of the proposed \textit{DIfferentiable Autoencoding Neural Operator (DIANO)} framework.  These scenarios are designed to evaluate DIANO across a range of spatio-temporal modeling tasks, with particular emphasis on coarse-grained latent-space visualization and physics-constrained latent evolution. Each configuration is described below:

\begin{enumerate}[label=(\roman*)]
    \item \textbf{Nonlinear Dimensionality Reduction (Static Mapping):}  
    This baseline setting focuses on spatial dimensionality reduction without temporal evolution. A high-dimensional input flow field at time $t^n$ is projected onto a coarse latent space and subsequently reconstructed at the same time instance.   
    \[
    \mathbf{u}(t^n) 
    \ \xrightarrow{\ \text{Encoder} \ } \ 
    \mathbf{z}(t^n) 
    \ \xrightarrow{\ \text{Decoder} \ } \ 
    \hat{\mathbf{u}}(t^n)
    \]

    \item \textbf{Nonlinear Dimensionality Reduction with Temporal Marching:}  
    This configuration extends the first one by incorporating temporal dynamics into the latent space. The encoded representation at time $t^n$ is advanced to $t^{n+1}$ using a differentiable PDE-based time integrator. The updated coarse latent state is then decoded to reconstruct the high-dimensional field at the future time step, enabling latent-space temporal modeling.  
    \[
    \mathbf{u}(t^n) 
    \ \xrightarrow{\ \text{Encoder} \ } \ 
    \mathbf{z}(t^n) 
    \ \xrightarrow{\ \text{PDE Evolution} \ } \ 
    \mathbf{z}(t^{n+1}) 
    \ \xrightarrow{\ \text{Decoder} \ } \ 
    \hat{\mathbf{u}}(t^{n+1})
    \]

    \item \textbf{Geometrical Reduction with Temporal Marching:}  
    This setting explores a geometrical space reduction scenario, where a high-dimensional field on the geometric space \(D_h\) (e.g., 2D or 3D) is compressed into a lower-dimensional latent geometric space \(D_\ell < D_h\) (e.g., 1D or 2D). The latent representation is then temporally evolved using the differentiable PDE solvers corresponding to the reduced space, and the future state is decoded back to the original geometrical space resolution. This scenario highlights the framework’s ability to preserve physically meaningful dynamics under substantial geometrical compression.  
    \[
    \mathbf{u}_{D_h}(t^n) 
    \ \xrightarrow{\ \text{Encoder} \ } \ 
    \mathbf{z}_{D_\ell}(t^n) 
    \ \xrightarrow{\ \text{PDE Evolution} \ } \ 
    \mathbf{z}_{D_\ell}(t^{n+1}) 
    \ \xrightarrow{\ \text{Decoder} \ } \ 
    \hat{\mathbf{u}}_{D_h}(t^{n+1})
    \]

    \item \textbf{Many-to-One Functional Mapping via Latent Fusion:}  
    The final scenario investigates multi-input functional mappings. A set of $m$ high-dimensional input fields at time $t^n$ is independently encoded into latent variables, which are then fused and mapped using a latent PDE solver to produce a single output representation. The decoder maps this to the corresponding high-dimensional prediction at the same instant $t^n$. This configuration mimics complex interactions such as flow superposition and coupled dynamics.  
    \[
    \left\{ \mathbf{u}^i(t^n) \right\}_{i=1}^m 
    \ \xrightarrow{\ \text{Encoder} \ } \ 
    \left\{ \mathbf{z}^i(t^n) \right\}_{i=1}^m 
    \ \xrightarrow{\ \text{PDE Mapping} \ } \ 
    \mathbf{p}(t^n) 
    \ \xrightarrow{\ \text{Decoder} \ } \ 
    \hat{\mathbf{P}}(t^n)
    \]

\end{enumerate}

\begin{figure}  
    \centering
    \includegraphics[width=0.85\textwidth,height=\textheight,keepaspectratio]{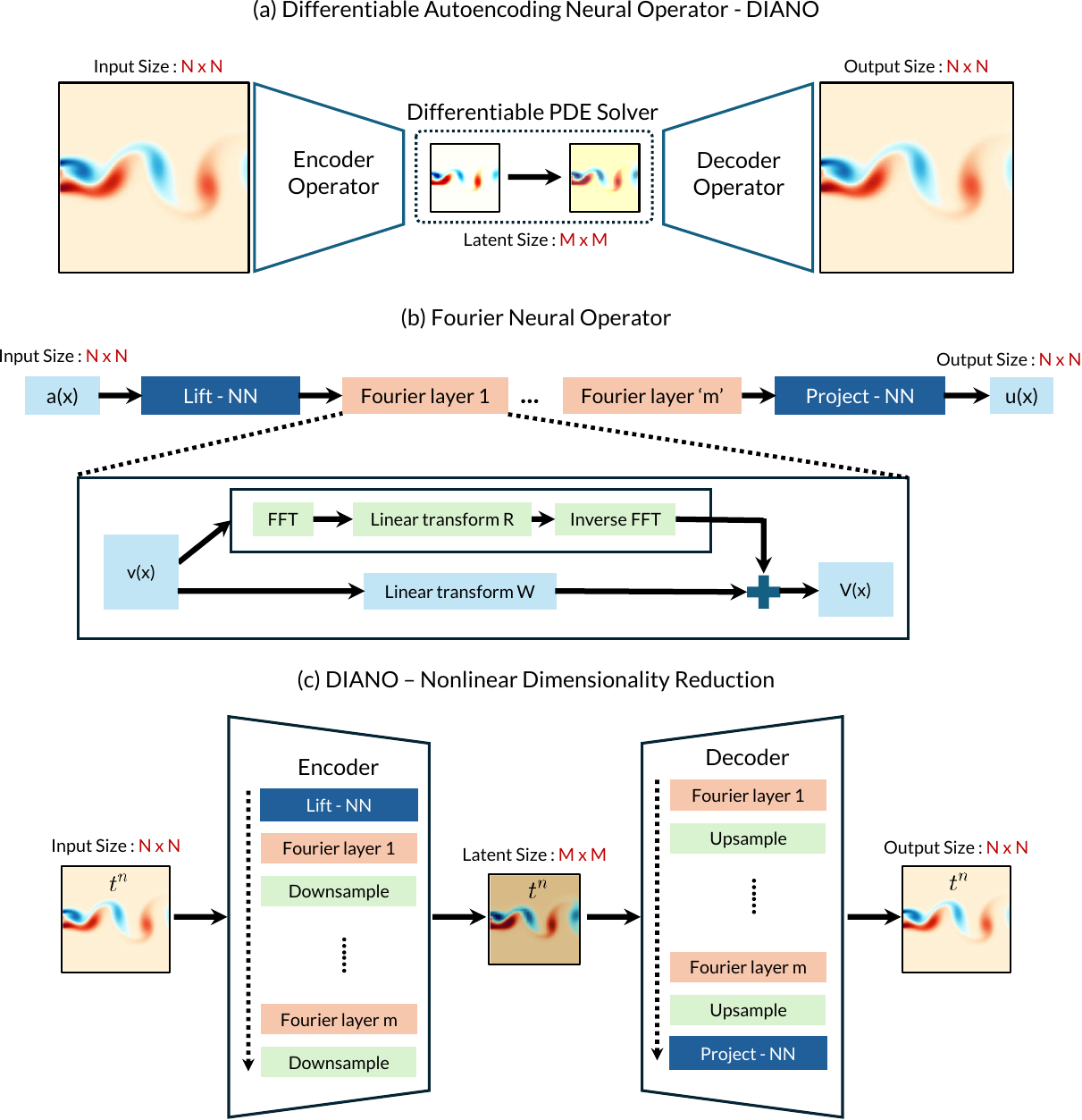}
\caption{
(a) Schematic of the proposed DIANO framework for spatiotemporal modeling. 
(b) Fourier Neural Operator (FNO), a neural operator using Fourier basis functions for spatial mapping.  
(c) Nonlinear dimensionality reduction (Static Mapping) with DIANO.
}

    \label{fig:diano}
\end{figure}

\subsection{Differentiable Autoencoding Neural Operator}
We introduce a differentiable autoencoding neural operator, DIANO, that integrates dimensionality/geometrical reduction, mesh-invariant functional mappings, a visualizable latent space, and physics-based temporal evolution into a unified framework. As illustrated in Fig.~\ref{fig:diano}a, the framework comprises three key components. First, a spatial encoder–decoder that compresses/coarsens the input field of size $N \times N$ into a compact latent representation of size $M \times M$, ($N > M$), while retaining essential information for accurate reconstruction. Second, functional basis layers that resolve spatial features within the autoencoder, enabling the efficient representation of input fields. A key innovation in this data reduction is that it acts as spatial coarsening, thereby producing field variables in the latent space that resemble the original field but on a coarser grid. Finally, a differentiable PDE solver is introduced that evolves the latent representation over time, enforcing physically consistent functional mapping and supporting end-to-end gradient-based optimization. Owing to the coarse grid structure of the latent space, differentiable PDE solvers can be defined thereon for the temporal evolution of such coarse-grained latent fields. In general, the differentiable PDE solver can be defined as a low-fidelity approximation of the original PDE, which is easier to solve.  Together, these components enable DIANO to learn complex spatiotemporal mappings while maintaining latent-space visualization and computational efficiency. 

\subsubsection{Autoencoding - Functional Basis Architecture}

Within the DIANO framework, spatiotemporal learning is explicitly decoupled, allowing the architecture to treat spatial representations and temporal dynamics independently. In contrast, when spatiotemporal modeling is ``coupled'', the choice of functional basis becomes critical and should reflect the underlying dynamics of the problem. Fourier bases are well-suited for periodic phenomena due to their global frequency decomposition, Laplace bases capture growth or decay processes with periodic modulation, and wavelet bases provide localized time–frequency resolution, making them effective for multi-resolution or transient dynamics. Such considerations have motivated the proliferation of neural operator variants tailored to specific classes of spatiotemporal problems in the literature. 

In DIANO, the spatial component is modeled following the FNO paradigm, with Fourier basis layers employed in both the encoder and decoder. As illustrated in Fig.~\ref{fig:diano}b, FNO consists of multiple Fourier layers, each composed of a Fourier transform, a spectral convolution, skip connections, and a nonlinear activation. Training retains low-frequency modes while truncating high-frequency ones to reduce computation and improve generalization. The output is obtained by transforming back to the physical domain. Our DIANO architecture builds on this FNO framework by incorporating explicit feature compression and decompression via convolutional operations (\textit{AvgPool2D} for downsampling and \textit{ConvTranspose2D} for upsampling), thereby developing a deterministic autoencoding neural operator. This architectural design enables the encoder to learn coarse-grained representations of the globally dominant structures in the latent space, while the decoder progressively corrects these latent representations into high-resolution, fine-grained outputs that effectively restore local structures, gradients, and variations, thereby retaining both local physical features and global spatial fidelity. Temporal evolution is handled in the latent space using a differentiable PDE solver. Unlike approaches that rely on explicit temporal learning operators, DIANO leverages this PDE solver to (i) enable physics-based time marching, (ii) provide a physically explainable characterization of latent temporal dynamics, and 
(iii) facilitate the incorporation of parametric variability (e.g., Reynolds number) in a more meaningful and efficient manner. This approach eliminates the need for data-driven temporal ML methods, while the autoencoder remains solely responsible for spatial compression (coarsening) and reconstruction of the flow fields. By decoupling temporal dynamics from spatial representation, this modular decomposition simplifies the learning task and allows for integration of various basis layers to resolve ``spatial'' features such as Laplace eigenfunctions, wavelet bases, Chebyshev polynomials, integral kernels, graph message passing, multipole expansions, attention weights, and kernel functions along with their associated neural operators. This flexibility facilitates the adaptation of the DIANO framework to the unique spatial characteristics of each problem, thus enhancing its applicability across a broad range of scenarios. In the present work, spatial representations are resolved using Fourier layers implemented within the FNO framework.

%If needed, the DIANO framework can incorporate alternative basis layers to construct spatial representations, adopting different neural operator architectures suited to specific spatial dynamics (e.g., Fourier, Laplace, or wavelet). This flexibility allows the DIANO framework to more accurately capture the dominant structures of the underlying spatial fields. However, it is important to note that the temporal component is consistently handled by the PDE solver.

\subsubsection{Differentiable PDE Solver}

Conventional numerical PDE solvers compute spatio-temporal solutions for given initial/boundary conditions and model parameters, offering high accuracy and stability for well-defined problems. However, such solvers do not construct a computational graph, which limits their compatibility with ML architectures and prevents direct computation of parameter gradients required for optimization. In contrast, differentiable PDE solvers compute the forward solution in the same manner as conventional solvers while simultaneously constructing a computational graph.  This structure enables automatic differentiation to obtain gradients with respect to any input parameter, without explicitly implementing an adjoint method. This capability enables efficient gradient-based optimization for physical parameters, boundary conditions, and other model inputs, allowing seamless integration into optimization and ML pipelines. The present framework leverages differentiable PDE solvers with the following key features:
\begin{itemize}
    \item \textbf{Latent-space embedding:} Placing the solver in a coarsened/compressed latent space enables efficient computation, significantly reducing the overall computational cost.  
    \item \textbf{Low-fidelity solver utilization:} The latent spatial field allows the use of a computationally inexpensive (lower fidelity) PDE solver that still captures essential dynamics.  
    \item \textbf{Latent Space–Flow Dynamics Correspondence:} The choice of PDE formulation, relative to the underlying governing dynamics, directly influences the coherency of latent coarse-grained features, as the solver has been utilized jointly during training.
\end{itemize}

The governing PDEs considered in this work are the 2D Vorticity Transport Equation (VTE) and the 3D Pressure Poisson Equation (PPE), which are applied to 2D flows (flow past a cylinder and flow through a symmetric stenosis) and 3D flows (patient-specific coronary blood flow), respectively. First, the 2D nonlinear VTE is given by
\begin{equation}
\frac{\partial \omega}{\partial t} 
+ u \frac{\partial \omega}{\partial x} 
+ v \frac{\partial \omega}{\partial y} 
= \nu \left( \frac{\partial^2 \omega}{\partial x^2} 
+ \frac{\partial^2 \omega}{\partial y^2} \right),
\label{vte}
\end{equation}
where $\omega$ is the vorticity, $u$ and $v$ are velocity components in the $x$ and $y$ directions, respectively, and $\nu$ is the kinematic viscosity. To investigate the impact of different PDE fidelity levels on the latent-space solution, several variants of the 2D VTE are considered:

\begin{enumerate}[label=(\roman*)]
    \item \textit{2D Linearized VTE:} Linearizes the convective terms assuming a characteristic uniform velocity scale $V$, capturing the balance between convection and diffusion of vorticity:
    \begin{equation}
    \frac{\partial \omega}{\partial t} + V \left( \frac{\partial \omega}{\partial x} + \frac{\partial \omega}{\partial y} \right) = \nu \left( \frac{\partial^2 \omega}{\partial x^2} + \frac{\partial^2 \omega}{\partial y^2} \right).
    \label{vte1}
    \end{equation}

    \item \textit{2D Stokes Flow (VTE without convection terms):} Neglects convective transport, corresponding to viscous-dominated Stokes flow for low Reynolds numbers:
    \begin{equation}
    \frac{\partial \omega}{\partial t} = \nu \left( \frac{\partial^2 \omega}{\partial x^2} + \frac{\partial^2 \omega}{\partial y^2} \right).
    \label{vte2}
    \end{equation}
    \item \textit{2D Inviscid Linearized VTE:} Removes diffusion terms to model inviscid flow, where vorticity is transported solely by convection:
    \begin{equation}
    \frac{\partial \omega}{\partial t} + V \left( \frac{\partial \omega}{\partial x} + \frac{\partial \omega}{\partial y} \right) = 0.
    \label{vte3}
    \end{equation}

    \item \textit{1D Linearized VTE along the streamwise direction ($x$):} Considers vorticity transport only along $x$, highlighting convection and diffusion along the primary flow:
    \begin{equation}
    \frac{\partial \omega}{\partial t} + V \frac{\partial \omega}{\partial x} = \nu \frac{\partial^2 \omega}{\partial x^2}.
    \label{vte4}
    \end{equation}

    \item \textit{1D Linearized VTE along the normal direction ($y$):} Considers vorticity transport only along $y$, isolating dynamics perpendicular to the main flow:
    \begin{equation}
    \frac{\partial \omega}{\partial t} + V \frac{\partial \omega}{\partial y} = \nu \frac{\partial^2 \omega}{\partial y^2}.
    \label{vte5}
    \end{equation}
\end{enumerate}

Secondly, the 3D Pressure Poisson Equation is given by
\begin{equation}
\frac{\partial^2 p}{\partial x^2} + \frac{\partial^2 p}{\partial y^2} + \frac{\partial^2 p}{\partial z^2} 
= -\rho \Bigg[
\left(\frac{\partial u}{\partial x}\right)^2 + 
\left(\frac{\partial v}{\partial y}\right)^2 + 
\left(\frac{\partial w}{\partial z}\right)^2
+ 2 {\left(\frac{\partial u}{\partial y}\frac{\partial v}{\partial x} + 
\frac{\partial u}{\partial z}\frac{\partial w}{\partial x} + 
\frac{\partial v}{\partial z}\frac{\partial w}{\partial y}\right)}
\Bigg],
\label{ppe}
\end{equation}
where $p$ is pressure, $(u, v, w)$ are velocity components in $(x, y, z)$, and $\rho$ is fluid density.

It should be noted that the VTE and its variants are parabolic PDEs, while the PPE is an elliptic PDE; consequently, the former require a time-marching direct solver, whereas the latter is solved using an iterative solver. Our differentiable solver is implemented using the \emph{finite difference method} (FDM) for both PDEs. The first derivatives are computed using an optimized upwind compact scheme (OUCS2)~\cite{sengupta2003analysis} to maintain the upwinding properties, while the second derivatives employ central difference schemes to preserve isotropy. For the VTE, time integration is performed using an explicit Runge-Kutta 4 (RK4) scheme, whereas for the PPE, the iterative Point-Jacobi method is used to compute the pressure field. The solver is fully differentiable, making it suitable for parameter estimation of $\nu$, $\rho$, and $V$, enabling latent PDE discovery and other gradient-based optimization tasks, when needed.

%Since inverse modeling with the DIANO framework is beyond the scope of the present work, we instead illustrate the efficacy of the PDE solver’s differentiability in identifying the governing latent PDE. As an illustrative example, we consider the 2D linearized VTE (Eq.~\ref{vte1}) and treat key parameters—viscosity and velocity scale—as trainable variables. The results, presented in~\ref{append2}, demonstrate how different training strategies for these parameters affect the interpretability of the recovered latent PDE with respect to the input function. This experiment serves as a proof-of-concept, highlighting DIANO’s potential for PDE model discovery. A more detailed and systematic investigation of inverse modeling and parameter training strategies is left for future work.

\begin{figure}  
    \centering
    \includegraphics[width=0.85\textwidth,height=\textheight,keepaspectratio]{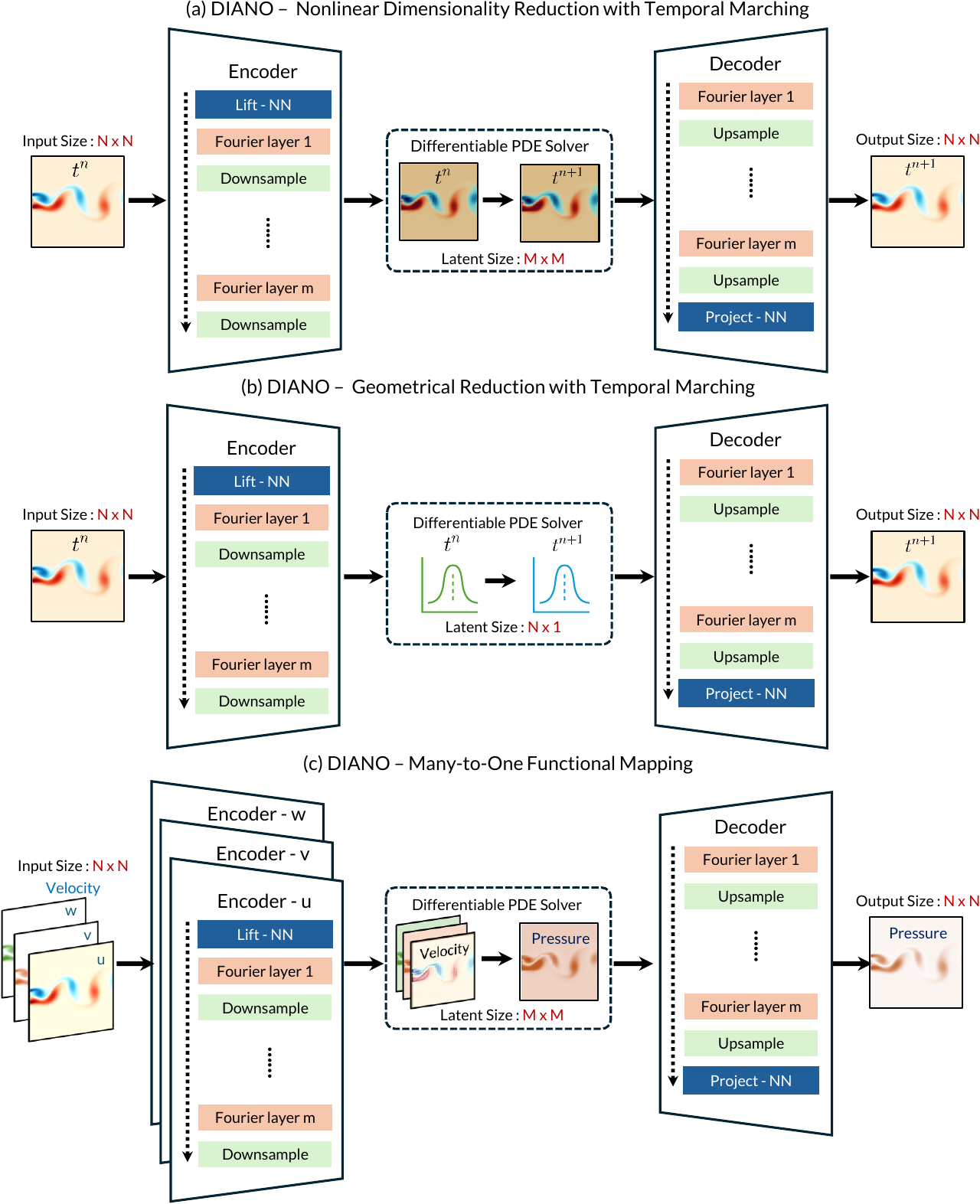}
    \caption{
DIANO architectural variants for each modeling scenario. The \emph{encoder} compresses inputs via spatial downsampling after each Fourier layer, and the \emph{decoder} reconstructs outputs via upsampling. Downsampling and upsampling are implemented with \textit{AvgPool2D} and \textit{ConvTranspose2D}, respectively.  
(a) Nonlinear Dimensionality Reduction with Temporal Marching: An unsteady PDE is solved in latent space from \( t^n \) to \( t^{n+1} \).  
(b) Geometrical Reduction with Temporal Marching: A 2D input is compressed to 1D, evolved via a given 1D PDE, and decoded back to 2D.  
(c) Many-to-One Functional Mapping: Three velocity components (u, v, and w) are independently encoded and the 3D PPE equation is solved in latent space to produce the latent pressure field, which is then decoded to full resolution.
}

\label{fig:diano2}
\end{figure}

\subsubsection{DIANO framework variants}
With the Fourier basis serving as the functional basis in the autoencoder along with the differentiable PDE solver, the DIANO framework encompasses multiple variants, each designed to address the four distinct objectives explained above. These architectural variants are described below.

\textit{Nonlinear Dimensionality Reduction - Static Mappings:} As shown in Fig.~\ref{fig:diano}c, DIANO is applied for dimensionality reduction, where no temporal evolution is required. In this sense, DIANO can be regarded as the deterministic counterpart of the Variational Autoencoding Neural Operator (VANO) framework~\cite{seidman2023variational}. The DIANO pipeline parallels FNO but introduces explicit compression and reconstruction mechanisms. In the encoder, inputs are first lifted to a higher-dimensional feature space via a neural lifting network (Lift-NN), subsequently processed through Fourier layers where low-frequency modes are preserved and high-frequency modes truncated, and finally compressed using \textit{AvgPool2D} by a factor of 2. Sequential application of Fourier transformations combined with downsampling defines the compression ratio and yields a compact, low-dimensional latent representation defined on a grid that could be visualized. In the decoder, this procedure is inverted, where latent features are passed through Fourier layers to recover spectral information and upsampled with \textit{ConvTranspose2D} by a factor of 2, progressively to restore spatial resolution. Sequential Fourier–upsampling steps, followed by a final projection through a neural projection network (Project-NN), reconstruct the output field at full resolution.

\textit{Nonlinear Dimensionality Reduction with Temporal Marching:}
In this temporal marching scenario, DIANO extends the static mapping variant by incorporating a differentiable PDE solver in the latent space, as depicted in Fig.~\ref{fig:diano2}a. The encoder compresses the spatial input features at time $t^n$ into a low-dimensional latent representation (a coarse grid), which is then evolved forward to $t^{n+1}$ using the PDE solver. The decoder reconstructs the spatial fields at $t^{n+1}$ from the evolved latent state, following the same Fourier–upsampling pipeline as in the static mapping case, thereby producing a high-dimensional output solution at the new time step. This variant offers three notable advantages. First, it alleviates the need for any additional machine learning-based temporal operator, instead leveraging the inherent structure of pre-defined governing equations for temporal evolution. Second, it enables systematic exploration of different PDE fidelities, such as simplified variants (e.g., simplified VTE formulations) of the governing PDEs used to generate the ground truth data directly within the latent space. This not only allows systematic investigation of how PDE model fidelity influences latent temporal evolution, but also contributes to explainability by establishing a direct correspondence between the latent dynamics and the underlying physics of the pre-defined differentiable PDE. Third, parametric variability, such as Reynolds number dependence, can be naturally incorporated through the latent PDE solver, facilitating the model to learn parameter-dependent dynamics in a physically meaningful and structured manner, which is found to be beneficial for generalization across both interpolation and extrapolation regimes.

\textit{Geometrical Reduction with Temporal Marching:}  
For the geometrical reduction case, DIANO follows the temporal marching approach described above but differs in the stage at which downsampling and upsampling are applied within the autoencoder, as illustrated in Fig.~\ref{fig:diano2}b. In contrast to the temporal variant, where the dimensionality of the 2D input data is evenly compressed and decompressed along both spatial dimensions during downsampling and upsampling, geometrical reduction performs downsampling and upsampling along only one dimension, leaving the other dimension unaltered. This results in encoding the 2D input into a 1D latent representation and decoding the 1D latent state back into 2D output data. Specifically, the two-dimensional input fields at time $t^n$ are first processed through sequential Fourier layers without reduction, and dimensionality compression is performed only along one direction at the final encoder stage. This produces a one-dimensional latent representation at $t^n$, which is then advanced to $t^{n+1}$ using the PDE solver. The evolved 1D latent state is subsequently restored to a 2D representation through a single upsampling step at the first stage of the decoder, after which sequential Fourier layers in the decoder refine spectral features and reconstruct the output solution at $t^{n+1}$. This architecture explicitly enforces a 2D$\rightarrow$1D$\rightarrow$2D geometrical reduction across the latent space, while maintaining temporally consistent evolution governed by the 1D PDE solver. Such an architecture enables one to learn geometric dimensionality reduction equipped with a geometrically reduced-order PDE model.

\textit{Many-to-One Functional Mapping via Latent Fusion:} In this case, DIANO is designed to represent many-to-one functional mappings, where multiple input fields are mapped to a single target output, as shown in Fig.~\ref{fig:diano2}c. Specifically, the full-resolution 3D  velocity components (u, v, and w) at time $t^n$ are independently encoded into low-dimensional latent spaces via three separate encoders, producing three 3D latent representations of the velocity field. These latent velocity representations are fused (stacked one after the other) and fed through the differentiable PPE solver to compute the corresponding low-dimensional latent representation of the pressure field. Finally, a single decoder reconstructs the full-resolution 3D pressure field at $t^n$ from this latent representation, thereby integrating information from multiple velocity components into a single pressure output. This architecture explicitly enables a many-to-one mapping in the latent space, defined as a coarse grid.

\subsection{Benchmark Flow Problems}
To evaluate the modeling capabilities of the proposed DIANO framework, three unsteady benchmark flow problems are considered. These include a canonical external flow configuration commonly employed in machine learning studies, as well as internal blood flow flows relevant to biomedical applications. The selected cases differ in geometric complexity and temporal boundary conditions, providing diverse and representative scenarios for assessing model robustness and for visualizing physical latent-space dynamics.

\begin{figure}  
    \centering
    \includegraphics[width=0.7\textwidth,height=\textheight,keepaspectratio]{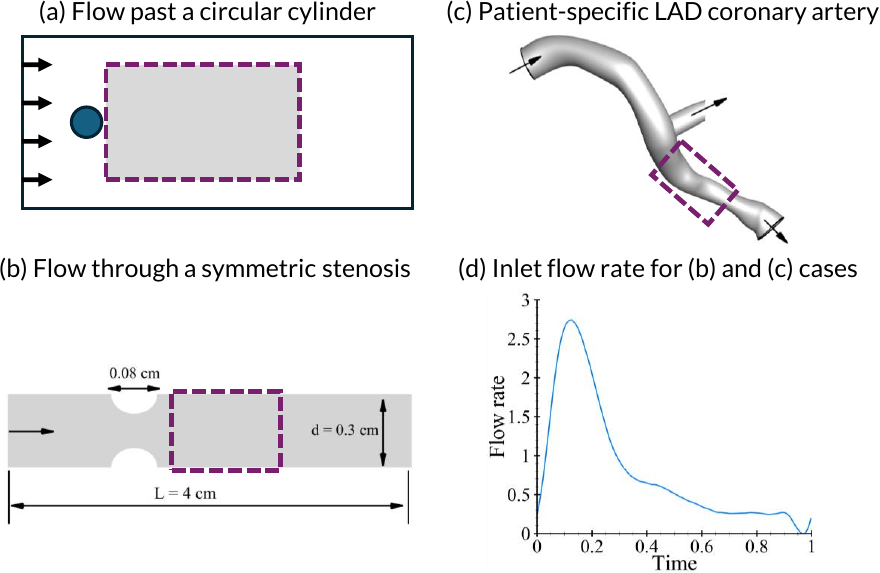}
    \caption{Benchmark flow problems considered in this study. (a) Flow past a 2D circular cylinder.  (b) Flow through an idealized, symmetric 2D arterial stenosis. (c) Flow through a patient-specific stenosed left anterior descending (LAD) coronary artery. (d) Transient inlet flow rate ($cm^3/s$) is employed for both cases (b) and (c). The region enclosed by the dotted box indicates the domain used for training and evaluating variants of the autoencoding architectures.}

    \label{fig:geometry}
\end{figure}

\textit{Flow over a Cylinder:} The first benchmark involves 2D incompressible flow over a circular cylinder at \( \mathrm{Re} = 100 \), a regime characterized by periodic vortex shedding. As shown in Fig.~\ref{fig:geometry}a, the cylinder (diameter \( D = 1 \)) is centered at the origin. The computational domain spans \( x \in [-10, 40] \) and \( y \in [-10, 10] \). No-slip boundary conditions are imposed on the cylinder, with uniform inflow \( (1, 0) \) and zero-traction outflow at the downstream boundary. The unsteady Navier–Stokes equations are solved using FEniCS~\cite{logg2012automated} with second-order triangular elements ($\sim$ 100K cells) and a time step of \( \Delta t = 0.01 \). After discarding transients, 1000 vorticity snapshots are extracted from \( t \in [100, 200) \), cropped to the wake region (indicated by a dotted box), normalized by their respective maximum vorticity values such that the resulting values lie in [-1, 1], and resampled onto a \( 256 \times 256 \) mesh. The resampling maps the physical domain to a normalized coordinate system, scaling the streamwise direction to [0, 1] and the transverse direction to [-1, 1]. This dataset, initially generated in prior work~\cite{csala2022comparing}, supports the first two modeling objectives of the present work: static and time-evolved nonlinear dimensionality reduction.

\textit{Pulsatile Flow through Stenosed Arteries:} The second and third benchmark problems involve pulsatile flow through a stenosed (blocked) coronary artery. The second case, shown in Fig.~\ref{fig:geometry}b, uses a 2D idealized symmetric stenosis with a 50\% blockage, 4~cm length, and 0.3~cm diameter. The third case, depicted in Fig.~\ref{fig:geometry}c, uses a patient-specific left anterior descending (LAD) coronary artery model containing an elongated soft plaque downstream of a bifurcation~\cite{mahmoudi2021story}. Both simulations use a physiologically realistic inflow waveform~\cite{kim2010patient} applied over five cardiac cycles, plotted in Fig.~\ref{fig:geometry}d, with a parabolic inlet velocity profile reaching peak \( \mathrm{Re}_{\text{max}} = 237 \) and mean \( \mathrm{Re} = 141 \). Only the final cardiac cycle is used for training and evaluation to exclude initial transients.

In the symmetric stenosis case, simulations are again performed in FEniCS with around 100K cells and \( \Delta t = 10^{-2} \). From the final cycale, 100 vorticity snapshots, focused on the post-stenotic recirculation zone (containing flow separation and reattachment) as denoted by a dotted rectangular box  are extracted, normalized to [-1, 1] by their maximum values, and resampled onto a \( 256 \times 256 \) mesh, mapping the streamwise direction to [0, 1] and the transverse direction to [-1, 1]. 
%A similar objective setup is explored in our prior work~\cite{csala2025physics} for improving 1D blood flow models with 3D data via neural PDEs, thus motivating the demonstration of DIANO's capability for the third objective-- geometrical reduction with temporal marching.

For the patient-specific LAD artery, transient simulations are conducted using SimVascular~\cite{updegrove2017simvascular} with a mesh of 3.7 million tetrahedral elements (1.4 million nodes), time step \( \Delta t = 5 \times 10^{-4} \)~s, and outlet resistances based on Murray's law (exponent 2.6, mean pressure 100 mmHg). The spatial domain is selected around the stenosis, indicated by a dotted box, where the dataset of 100 snapshots of containing 3D velocity and pressure fields are extracted, normalized by their respective maximum values—velocity components and pressure scaled to [0, 1], and resampled onto a structured \( 256 \times 256 \times 256 \) grid with coordinates normalized to [0, 1] in all directions. Mapping the 3D arterial data onto a structured grid introduced ghost points outside the vessel walls, which were assigned zero values. This dataset is used to demonstrate pressure inference from 3D velocity components via the Pressure Poisson equation, supporting the final objective: many-to-one functional mapping through latent fusion.

It is important to note that all three cropped ground truth datasets exhibit inherent temporal periodicity. In the case of flow past a cylinder, this periodicity arises from vortex shedding despite a steady inflow, whereas in arterial flow scenarios, it originates from the pulsatile nature of the inflow waveform. 
%This recurring temporal behavior naturally motivates the use of Fourier-based layers for temporal learning, owing to their efficiency in representing periodic functions through spectral decomposition. In contrast, the present DIANO framework adopts a fundamentally different strategy: temporal evolution is modeled through the differentiable PDE solver operating within the latent space, while spatial representations are resolved using Fourier basis layers corresponding to the FNO methodology.

\subsection{Experimental Setup}

\renewcommand{\arraystretch}{1.2}
\begin{sidewaystable*}
\centering
\begin{adjustbox}{width=\linewidth}
\begin{tabular}{|c|c|c|c|c|c|c|c|c|c|}
\hline
\multicolumn{10}{|c|}{\textbf{Nonlinear Dimensionality Reduction - Static Mapping}} \\ \hline
Methods & Epoch & Batch size & Learning rate & Step epoch & Decay rate & Parameters (M) & Train error & Test error & Training Time (hrs) \\ \hline
CNN-AE - CR=4 & 150 & 30 & $10^{-1}$ & 20 & 0.5 & 7.6 & $1.629\times 10^{-5}$ & $1.154\times 10^{-4}$ & 1.14 \\ \hline
CNO    & 70  & 30 & $10^{-2}$ & 5  & 0.75 & 31 & $3.669\times 10^{-6}$ & $6.752\times 10^{-6}$ & 0.27 \\ \hline
NN-AE - 8 LM  
& \multirow{3}{*}{2000} & \multirow{3}{*}{256} & \multirow{3}{*}{$10^{-2}$} 
& \multirow{2}{*}{100} & \multirow{2}{*}{0.5} & \multirow{3}{*}{273}
& $1.077 \times 10^{-5}$ & $1.220 \times 10^{-5}$ & \multirow{2}{*}{$0.13\pm 0.01$} \\ \cline{1-1} \cline{8-9}
NN-AE - 16 LM 
&  &  &  &  &  & 
& $8.448 \times 10^{-6}$ & $9.111 \times 10^{-6}$ &  \\ \cline{1-1} \cline{5-6} \cline{8-10}
NN-AE - 32 LM 
&  &  &  & 120 & 0.75 & 
& $7.267 \times 10^{-6}$ & $8.846 \times 10^{-6}$ & 0.14 \\ \hline
DIANO - FM=8, CR=16 
& \multirow{3}{*}{100} & \multirow{3}{*}{30} & \multirow{3}{*}{$10^{-2}$} 
& \multirow{3}{*}{5} & \multirow{3}{*}{0.75} & \multirow{3}{*}{1.2}
& $3.623\times 10^{-6}$ & $3.687\times 10^{-6}$ & \multirow{2}{*}{$0.17\pm 0.01$} \\ \cline{1-1} \cline{8-9}
DIANO - FM=16, CR=16 
&  &  &  &  &  & 
& $2.042\times 10^{-6}$ & $2.070\times 10^{-6}$ &  \\ \cline{1-1} \cline{8-10}
DIANO - FM=32, CR=16 
&  &  &  &  &  & 
& $3.907\times 10^{-7}$ & $3.944\times 10^{-7}$ & 0.18 \\ \hline
DIANO - FM=8, CR=4  
& \multirow{4}{*}{100} & \multirow{4}{*}{30} & \multirow{4}{*}{$10^{-2}$} 
& \multirow{4}{*}{5} & \multirow{4}{*}{0.75} & \multirow{4}{*}{1.2}
& $3.488\times 10^{-6}$ & $3.506\times 10^{-6}$ & 0.14 \\ \cline{1-1} \cline{8-10}
DIANO - FM=8, CR=16  
&  &  &  &  &  & 
& $3.623\times 10^{-6}$ & $3.687\times 10^{-6}$ & 0.17  \\ \cline{1-1} \cline{8-10}
DIANO - FM=8, CR=64  
&  &  &  &  &  & 
& $4.187\times 10^{-6}$ & $4.237\times 10^{-6}$ & 0.18 \\ \cline{1-1} \cline{8-10}
DIANO - FM=8, CR=256 
&  &  &  &  &  & 
& $5.279\times 10^{-6}$ & $5.334\times 10^{-6}$ & 0.20 \\ \hline
\multicolumn{10}{|c|}{\textbf{Nonlinear Dimensionality Reduction with Temporal Marching}} \\ \hline
2D Linear VTE          
& \multirow{5}{*}{60} & \multirow{5}{*}{30} & \multirow{5}{*}{$10^{-2}$} 
& \multirow{5}{*}{5} & \multirow{5}{*}{0.75} & \multirow{5}{*}{1.3}
& $3.219\times 10^{-6}$ & $2.661\times 10^{-6}$ & \multirow{3}{*}{$0.24 \pm 0.01$} \\ \cline{1-1} \cline{8-9}
2D Stokes flow - VTE   
&  &  &  &  &  & 
& $4.708\times 10^{-6}$ & $4.743\times 10^{-6}$ &  \\ \cline{1-1} \cline{8-9}
2D Inviscid Linear VTE 
&  &  &  &  &  & 
& $3.007\times 10^{-6}$ & $3.050\times 10^{-6}$ &  \\ \cline{1-1} \cline{8-10}
1D Linear VTE (x)      
&  &  &  &  &  & 
& $3.811\times 10^{-6}$ & $4.006\times 10^{-6}$ & \multirow{2}{*}{0.22} \\ \cline{1-1} \cline{8-9}
1D Linear VTE (y)      
&  &  &  &  &  & 
& $9.612\times 10^{-6}$ & $9.872\times 10^{-6}$ &  \\ \hline
\multicolumn{10}{|c|}{\textbf{Geometric Reduction with Temporal Marching}} \\ \hline
1D Linear VTE - x 
& \multirow{2}{*}{150} & \multirow{2}{*}{20} & \multirow{2}{*}{$10^{-2}$} 
& \multirow{2}{*}{10} & \multirow{2}{*}{0.75} & \multirow{2}{*}{0.38}
& $7.246\times 10^{-5}$ & $1.003\times 10^{-4}$ & \multirow{2}{*}{0.04} \\ \cline{1-1} \cline{8-9}
1D Linear VTE - y 
&  &  &  &  &  & 
& $3.088\times 10^{-5}$ & $4.743\times 10^{-5}$ &  \\ \hline
\multicolumn{10}{|c|}{\textbf{Many-to-One Functional Mapping}} \\ \hline
3D PPE                
& 800 & \multirow{2}{*}{2} & \multirow{2}{*}{$10^{-2}$} 
& 30 & \multirow{2}{*}{0.75} & \multirow{2}{*}{8}
& $5.260\times 10^{-5}$ & $3.70\times 10^{-3}$ & 3.33 \\ \cline{1-2} \cline{5-5} \cline{8-10}
3D PPE w/o $\nabla^2$ 
& 500 &  &  & 20 &  & 
& $9.280\times 10^{-5}$ & $4.06\times 10^{-3}$ & 2.41 \\ \hline
\end{tabular}
\end{adjustbox}
\caption{Hyperparameters of all methodologies considered for the four modeling objectives. Trainable parameters are reported in \textit{millions (M)}. For the static mapping, experiments compare NN-AE (varying the number of latent modes (LM)), CNO, CNN-AE, and DIANO with varying Fourier modes (FM) and compression ratios (CR). For the remaining modeling scenarios, only DIANO is employed to investigate its sensitivity to coarse-grained latent representations. Reported errors are the mean squared error (MSE). Training wall-clock time is measured on an NVIDIA RTX A4500 GPU.}
\label{tab:hyper1}
\end{sidewaystable*}

For both static mapping and temporal evolution scenarios, we employ loss functions that quantify the reconstruction error. Let the input flow field be $\mathbf{x}_i^n \in \mathcal{X}$, where $i$ indexes the training samples and $n$ represents the time step. Consider two different autoencoder architectures: convolution-based (CNN) and neural network-based (NN) autoencoders. The input space is defined as  

\[
\mathcal{X} =
\begin{cases}
\mathbb{R}^{H \times W \times C}, & \text{for CNN},\\
\mathbb{R}^{S}, \quad S = H \cdot W \cdot C, & \text{for NN},
\end{cases}
\]
where $H$, $W$, and $C$ denote the height, width, and number of channels. In the CNN-type architecture, the input $\mathbf{x}_i^n$ is processed as a spatial tensor, allowing the network to capture both local and global structures. The encoder $\mathcal{E}$ maps the input to a latent representation $\mathbf{z}_i \in \mathbb{R}^{H_z \times W_z \times C_z}$, and the decoder $\mathcal{D}$ reconstructs the field as $\hat{\mathbf{x}}_i^n = \mathcal{D}(\mathcal{E}(\mathbf{x}_i^n)) \in \mathbb{R}^{H \times W \times C}$. For the NN-based architecture, the input is first flattened into a vector $\mathbf{x}_{i, \rm flat}^n$, which discards explicit spatial structure. The encoder maps this vector to a lower-dimensional latent space $\mathbf{z}_i \in \mathbb{R}^{d}$, with $d \ll S$, and the decoder reconstructs the flow field as $\hat{\mathbf{x}}_i^n = \mathcal{D}(\mathcal{E}(\mathbf{x}_{i, \rm flat}^n)) \in \mathbb{R}^{S}$. 

The compression ratio (CR) quantifies the dimensionality reduction achieved by an autoencoder. For an input flow field defined on an $N \times N$ grid with a single channel, an NN-based autoencoder with latent dimension $d$ yields a compression ratio of $\mathrm{CR}_{\mathrm{NN}} = N^2 / d$. In contrast, for a CNN- or FNO-based autoencoder with a latent representation of size $M \times M$, it becomes $\mathrm{CR}_{\mathrm{CNN}} = N^2 / M^2$. This distinction reflects the different compression mechanisms of the two architectural variants: NN-based models reduce dimensionality globally, while CNN/FNO-based models retain spatial structure in the latent space.

In the static mapping scenario, models are trained to minimize the mean squared reconstruction error at the same time step,
\begin{equation}
\mathcal{L}_{\rm rec} = \frac{1}{N} \sum_{i=1}^N \|\mathbf{x}_i^n - \hat{\mathbf{x}}_i^n\|_2^2 \;,
\label{loss1}
\end{equation}
where $N$ is the number of training samples. In the many-to-one functional mapping setting, both $\mathbf{x}$ and $\hat{\mathbf{x}}$ correspond to the pressure field at a given time $t^n$. For temporal evolution scenarios, the objective is to predict the flow field at the next time step, and the reconstruction loss is replaced with a temporal prediction loss
\begin{equation}
\mathcal{L}_{\rm temporal} = \frac{1}{N} \sum_{i=1}^N \|\mathbf{x}_i^{n+1} - \hat{\mathbf{x}}_i^{n+1}\|_2^2 \;.
\label{loss2}
\end{equation}

In the present work, all architectures are trained using the ADAM optimizer, with parameter updates governed by the respective loss functions. A step-decay dynamic learning rate schedule is employed to promote systematic loss convergence of the training process. The hyperparameters utilized in each experiment are reported in Table~\ref{tab:hyper1}, along with the corresponding reconstruction performance on both the training and testing datasets. The ground truth datasets are randomly partitioned into $80\%$ for training and $20\%$ for testing, with manual seeding applied to ensure reproducibility across all architectural autoencoding variants presented herein. 

%Importantly, the ground truth datasets are normalized using the following formulations:
%\begin{equation}
%X = \frac{x - x_{\min}}{x_{\max} - x_{\min}}, \quad X \in [0, 1],
%\end{equation}

%\begin{equation}
%X = \frac{x - x_{\min}}{x_{\max} - x_{\min}} + \frac{x - x_{\max}}{x_{\max} - x_{\min}}, \quad X \in %[-1, 1].
%\end{equation}
%Because these transformations are based on known minimum and maximum values, it is straightforward to map the normalized variables back to their corresponding physical scales whenever necessary. In addition to the data normalization, this procedure is also useful for visualization, as it enables a clearer interpretation of the latent flow fields by effectively highlighting the spatial flow structures. In the DIANO framework, the encoder's output is not explicitly constrained by any activation function that would bound the latent space flow scales. It was found that applying any explicit bounding/clipping process comprises latent interpretability, even though reconstruction accuracy remains unaffected, as observed across all experiments conducted in this work.

\section{Numerical Experiments}
\label{sec4}
We evaluate the DIANO framework through a series of numerical experiments designed to assess its capability to learn compact and explainable latent representations while accurately capturing complex spatial and temporal dynamics. Our study focuses on four key aspects: (i) nonlinear dimensionality reduction, (ii) temporal evolution in latent space, (iii) geometrical compression, and (iv) many-to-one functional mappings. We begin by benchmarking DIANO against established autoencoding architectures to evaluate compression efficiency and latent space characteristics. Building upon this, we investigate the DIANO method on the remaining three scenarios.

\subsection{Nonlinear Dimensionality Reduction -- Static Mappings}

\begin{figure}  
    \centering
    \includegraphics[width=.9\textwidth,height=\textheight,keepaspectratio]{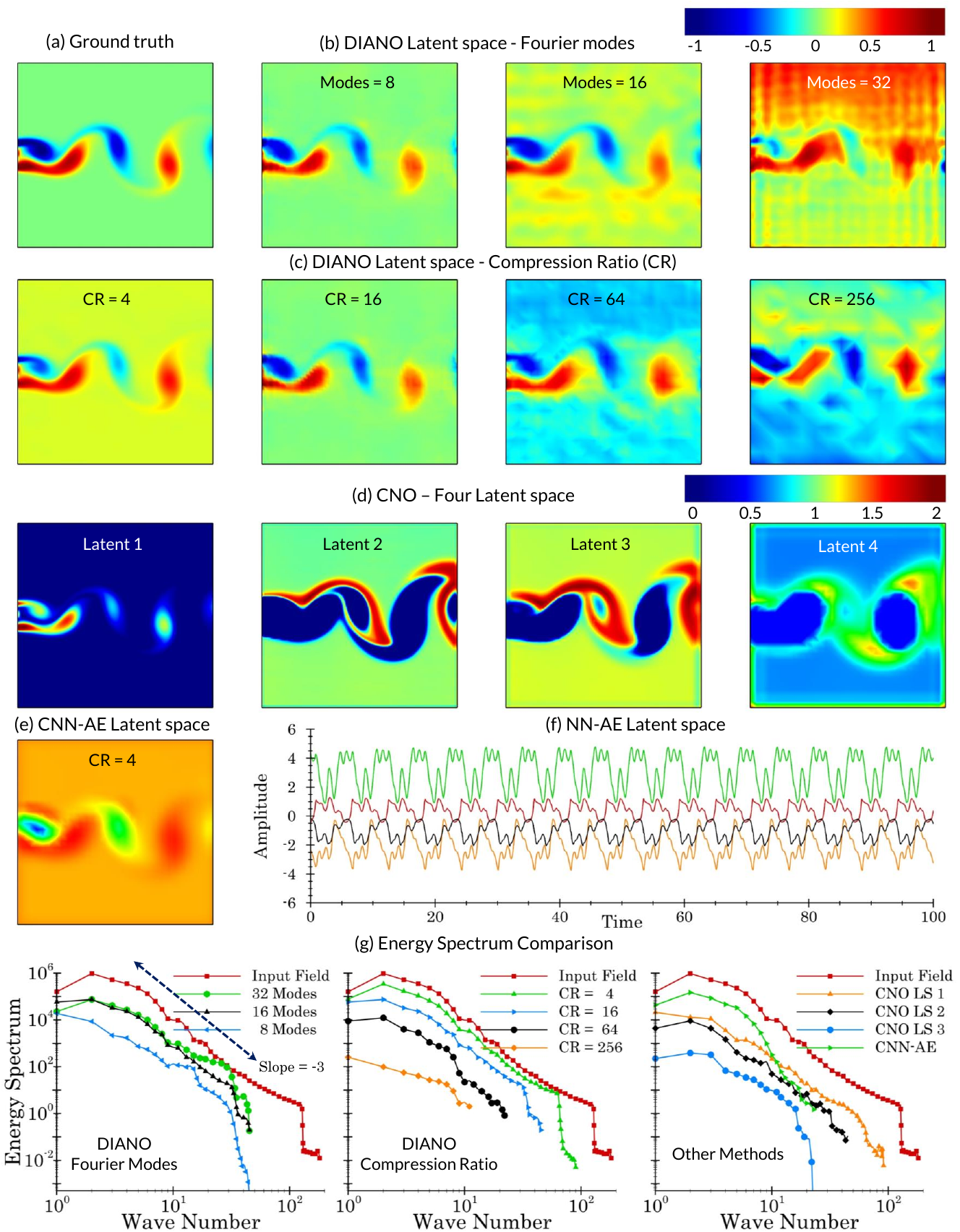}
    \caption{Nonlinear dimensionality reduction - Static Mapping. Comparison of the vorticity latent space structure across different autoencoding methodologies: CNO, CNN-AE, NN-AE, and DIANO. (a) Ground-truth vorticity field from the \textit{test dataset}, showing asymmetric vortex shedding; (b) DIANO: effect of the number of Fourier modes at compression ratio $=16$; (c) DIANO: effect of compression ratio with 8 Fourier modes; \textbf{(d)} CNO: latent representations at four compression levels;(e) CNN-AE: latent structure at compression ratio $=4$, using the panel b color scale; (f) NN-AE: temporal evolution of four randomly selected bottleneck modes; (g) Energy cascade spectrum of the vorticity field, comparing latent representations across Fourier modes, compression ratios, and other methods with the ground-truth.}
    \label{fig:dim_red_cylinder}
\end{figure}

The vorticity field of the von Kármán vortex street, exhibiting asymmetric shedding behind a cylinder, is employed to evaluate the reconstruction performance and, importantly, the latent coarse-grained structures of DIANO across varying compression ratios, where the inputs and outputs correspond to the same time instant $t^n$. To provide a benchmark, DIANO is compared with established autoencoding architectures, including convolutional autoencoders (CNN-AE), fully connected autoencoders (NN-AE), and CNO. Detailed implementation of these baseline methods is provided in~\ref{append1}.

Experiments with NN-AE are performed by varying the number of latent modes (8, 16, and 32) to investigate reconstruction accuracy under different levels of input compression. In contrast, DIANO, which incorporates a Fourier basis layer, introduces two key hyperparameters: (i) the number of Fourier modes resolved within a Fourier layer, and (ii) the number of Fourier layers with upsampling/downsampling, which controls the compression of the input data to the latent space grid size. For CNN-AE, a fixed compression factor of 4 is applied to examine latent representations. Notably, CNO utilizes a U-Net-based autoencoding approach that produces latent representations at multiple compression levels in a single forward pass. Additionally, all latent features in CNO have access to high-dimensional information through residual blocks, which fundamentally mimic multigrid solvers.

The latent space vorticity structures for each method are presented in Figure~\ref{fig:dim_red_cylinder}, with the structures constructed based on the ground-truth vorticity field from the test dataset, which exhibits asymmetric vortex shedding (Fig.~\ref{fig:dim_red_cylinder}a). For DIANO, the number of Fourier modes within a Fourier layer is varied at a fixed compression ratio of 16 (two Fourier layer-upsampling blocks in the encoder and two Fourier layers-downsampling blocks in the decoder) to investigate the sensitivity of Fourier modes to the latent structure. The results, shown in Fig.~\ref{fig:dim_red_cylinder}b, indicate that 8 Fourier modes sufficiently capture the dominant vortical structures of the given input data, producing a clean and physically meaningful coherent latent representation.  As the number of Fourier modes increases, higher-frequency components are learned, introducing fine-scale features that interfere with the coarse-grained coherent vortices. This effect is particularly noticeable for the high-mode case (\(32\) Fourier modes), where the latent representation becomes less distinct and apparent. Therefore, using the minimal number of Fourier modes necessary to resolve the dominant spatial features yields a clearer and more physically meaningful latent structure.  Nevertheless, the reconstruction error reported in Table~\ref{tab:hyper1} exhibits an opposite trend: reconstruction errors decrease from \(\mathcal{O}(10^{-6})\) for low Fourier mode configurations to \(\mathcal{O}(10^{-7})\) for high-mode configurations, consistent with the physical expectation that higher-frequency modes should resolve much finer scales of vorticity dynamics. This behavior highlights a fundamental trade-off: fewer Fourier modes promote distinct, coherent coarse-grained latent vortical structures at the expense of reconstruction accuracy, whereas a higher mode count improves reconstruction fidelity at the cost of latent structures' coherency. Therefore, selecting a moderate number of Fourier modes offers a balanced compromise, achieving sufficient reconstruction accuracy while preserving coherent latent structures for analysis.

For a fixed number of Fourier modes of 8, the compression ratio (CR) of the DIANO framework is systematically varied by increasing both the number of Fourier layer-downsampling blocks and Fourier layer-upsampling blocks simultaneously in the encoder and decoder, respectively. This approach is used to assess the coherency of the latent vortical structures under different levels of compression, and the results are depicted in Fig.~\ref{fig:dim_red_cylinder}c. As anticipated, the latent vorticity structures become progressively coarser as the compression ratio increases. In other words, the coherence and spatial resolution of the learned latent vortical structures decrease when the compression gets stronger, leading to a less detailed representation of the flow features in the latent space. In contrast to the previous scenario regarding the Fourier modes used, coarsening leads to a consistent increase in reconstruction error with increasing compression ratio, as reported in Table~\ref{tab:hyper1}. 

In comparison, CNO generates hierarchical latent representations across multiple compression levels, and the extracted latent structures at increasing compression levels are presented in Fig.~\ref{fig:dim_red_cylinder}d. Notably, while the latent representations across these compression levels are physically explainable, several limitations arise when compared with the asymmetric vortex street in the ground-truth data. In the first latent space (Latent 1), corresponding to the input dimension (i.e., no compression), vortical structures are identifiable; however, directional information is lost, preventing distinction between clockwise and anticlockwise vortices, a key feature of the asymmetric von Kármán vortex street. In the subsequent two latent spaces, with compression ratios of 4 and 16, further vortical structure information is lost, although the contours of regions containing vortical structures remain partially observable. In the final latent space (Latent 4), at a compression ratio of 64, essentially no physically relevant information is retained, with only some flow structures aligned along the streamwise direction, offering minimal insight into the underlying flow dynamics. Thus, while CNO produces hierarchical latent representations in a single forward pass, its latent spaces are relatively less physically meaningful compared to DIANO, which provides significantly better meaningful structures, a single latent representation (instead of multiple ones in CNO), with fewer trainable parameters and comparable reconstruction accuracy, as shown in Table~\ref{tab:hyper1}.

Lastly, the latent structures of the CNN-AE and NN-AE autoencoders are presented in Figs.~\ref{fig:dim_red_cylinder}e and~\ref{fig:dim_red_cylinder}f, respectively. For the CNN-AE case, a naive architecture produces latent representations that lack physical meaningfulness and fail to capture coherent vortical structures (not shown), whereas introducing residual connections (discussed in~\ref{append1}), similar to the CNO architecture, yields a more structured latent space in which coherent vortical patterns are recovered with a clear correspondence to the input field. In the NN-AE case, the latent space is inherently unexplainable over space, across three chosen latent modes, although a good reconstruction error of $\mathcal{O}(10^{-5})$ is achieved. Nevertheless, the latent dynamics exhibit periodic temporal behavior with multiple harmonics, consistent with the ground-truth vortex shedding. While such spatial behavior is expected, temporal evolution can be learned by modeling the latent dynamics via ODE discovery. This is demonstrated by coupling an autoencoder with Sparse Identification of Nonlinear Dynamics (SINDy)~\cite{champion2019data}, which incorporates latent dynamics into training, while LaSDI~\cite{fries2022lasdi} follows a two-step approach using LASSO to identify the governing ODE after training the autoencoder.

Building on the qualitative observations above, we now assess how well the coarse-grained latent representation is aligned with the high-resolution dynamics in a \textit{quantitative} sense. The results, summarized in Fig.~\ref{fig:dim_red_cylinder}g, are reported across variations in Fourier modes, compression ratios, and model architectures. We employ the vorticity energy (enstrophy) spectrum across wavenumber ($k$) as a diagnostic metric. For the present flow configuration, the inertial-range energy cascade is expected to follow a $k^{-3}$ scaling~\cite{singh2004energy}. The reference ground-truth spectrum is computed on a structured $256 \times 256$ grid used for training the ML models, interpolated from the original unstructured data. The reference $k^{-3}$ scaling, corresponding to the input vorticity field, is indicated by a dotted line and compared with latent vorticity fields obtained from DIANO for different Fourier modes at a fixed compression ratio of 16. Across all configurations, energy spectra of all latent vorticity fields exhibit a decay close to the expected $-3$ slope. Furthermore, as the number of Fourier modes (i.e., retained frequencies in each Fourier layer) increases, the magnitude of the latent spectrum progressively approaches the ground-truth spectrum. This indicates that the number of Fourier modes can serve as an effective hyperparameter for controlling agreement with the ground truth. For a fixed Fourier modes of 8, compression ratios of 4, 16, and 64 yield latent spectra that nearly follow the reference scaling over the resolved wavenumber range, which is directly constrained by the level of compression. As the compression ratio increases, high-wavenumber components are progressively truncated, reducing spectral resolution. At a compression ratio of 256, this truncation becomes severe, leading to significant loss of high-wavenumber content and limiting the representation of inertial-range dynamics.  The vorticity energy spectra obtained from CNO and CNN-AE are also compared against the ground-truth spectrum (right-most panel). Notably, although the CNO latent space lacks directional coherence in representing vortex shedding structures, it nonetheless preserves quantitative consistency with the ground-truth energy spectrum across its hierarchical latent representations. In contrast, the CNN-AE with residual connections, while producing qualitatively explainable latent structures, fails to reproduce the expected  $k^{-3}$ energy decay, indicating that qualitative structures would not always guarantee quantitative physical fidelity.

These results indicate that DIANO achieves a favorable balance between compression and the coherency of latent vortical structures, effectively capturing essential flow features while maintaining a compact latent representation. From a qualitative perspective, DIANO preserves coherent vortical structures in the latent space, while quantitatively, it maintains consistency with the underlying energy cascade behavior across the experiments considered. This balance is determined by the choice of Fourier modes and the compression ratio, which control the trade-off among reconstruction fidelity, structural resolution, and spectral accuracy.

\subsection{Nonlinear Dimensionality Reduction with Temporal Marching}\label{sec3}
\subsubsection{Latent Coherent Structures and Solver Fidelity: Prediction Accuracy}
This section presents the prediction of future temporal solutions and the latent characteristics of the DIANO framework. The focus is on the asymmetric von Kármán vortex street generated downstream of a circular cylinder, a canonical benchmark for unsteady wake dynamics. The prediction task is formulated as a one-step temporal advancement problem, where the vorticity field at time $t^n$ serves as input and the corresponding solution at $t^{n+1}$ is obtained. The DIANO framework employs a differentiable PDE solver within the latent space, designed to evolve the Vorticity Transport Equation (VTE). The solver acting on the latent representation enables the temporal advancement to remain consistent with the governing dynamics in a meaningful fashion.  

\begin{figure}  
    \centering
    \includegraphics[width=0.8\textwidth,height=\textheight,keepaspectratio]{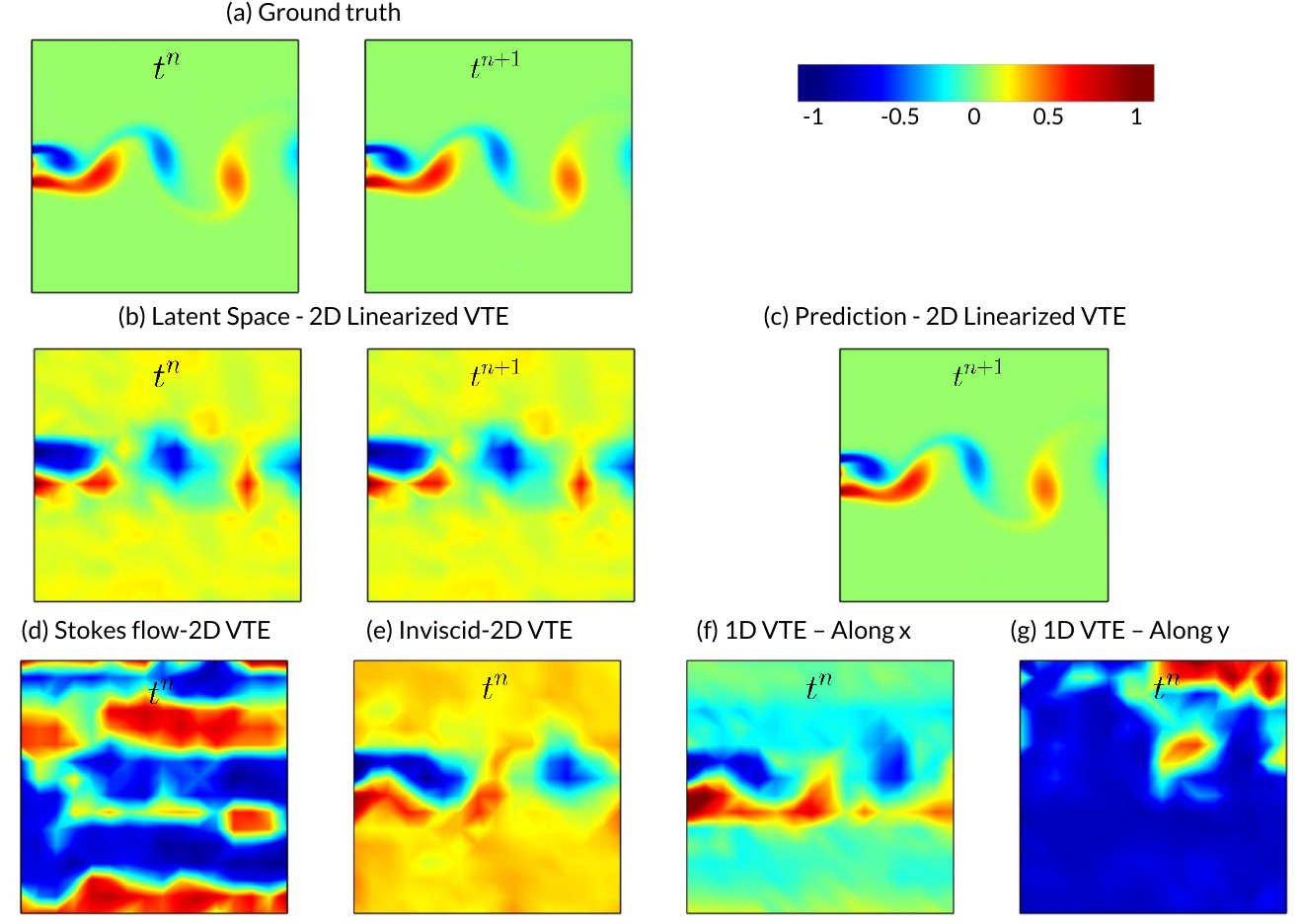}
    \caption{Nonlinear Dimensionality Reduction with Temporal marching. Comparison of vorticity latent space structure of the DIANO framework  (compression ratio 256, 16 Fourier modes)  across different VTE simplified formulations solved within the latent space. The vorticity field of the von Kármán vortex street behind a cylinder is used as the benchmark case. (a) Ground truth data from the \textit{test} dataset, where the input and output flow fields correspond to time instants $t^{n}$ and $t^{n+1}$, respectively. (b) Latent-space vortical structures at $t^{n}$ and $t^{n+1}$ computed by the PDE solver, and (c) the corresponding predicted output flow field at $t^{n+1}$ for 2D linearized VTE. Latent-space vortical structures at $t^{n}$ obtained using: (d) 2D Stokes flow (linearized VTE without the convection term), (e) 2D inviscid linearized VTE, (f) 1D linearized VTE along the streamwise direction $x$, and (g) 1D linearized VTE along the normal direction $y$.}
    \label{fig:temp_march_cylinder}
\end{figure}

The results of temporally marched latent vortical structures obtained using the DIANO framework employing 16 Fourier modes in each Fourier layer of the autoencoder and a compression ratio of 256 are presented in Fig.~\ref{fig:temp_march_cylinder}. These results are based on the ground truth flow fields from the \textit{test} dataset at time instants $t^{n}$ and $t^{n+1}$, showing the asymmetric von Kármán vortex street downstream of the cylinder, as illustrated in Fig.~\ref{fig:temp_march_cylinder}a. Despite the ground truth data being generated from the fully coupled 2D nonlinear Navier-Stokes equations, the DIANO framework attempts to advance the flow in the latent space using simplified variants of the VTE (Eqs.~\ref{vte1}-~\ref{vte5}). As demonstrated, the concurrent training of the compression together with the simplified VTE equation enforces the smaller flow scales and fine features to be filtered out by the encoder, leaving only the dominant flow structures in the latent representation. Upon evolution of the latent vorticity solution to a future time $t^{n+1}$ using the PDE solver, the decoder now reconstructs the fine-scale features that were filtered by the encoder, thereby recovering an effective approximation of the full flow field. The differentiable VTE solver uses a characteristic velocity scale $V = 1$ and kinematic viscosity $\nu = 0.01$, chosen based on the Reynolds number. Time integration is performed with a time step of $0.01$ for all cases presented in this section.  It is imperative to note that the ground truth flow is advection-dominated, with advection stronger along the horizontal streamwise direction (\textit{x}) relative to the vertical normal direction (\textit{y}).

In Fig~\ref{fig:temp_march_cylinder}, DIANO operating with the 2D linearized VTE (Eq.~\ref{vte1}) is presented, showing the latent vorticity fields at $t^n$ and the temporally marched solution at $t^{n+1}$ in Fig.~\ref{fig:temp_march_cylinder}b, along with the corresponding predicted output vorticity field at $t^{n+1}$ in Fig.~\ref{fig:temp_march_cylinder}c. With a strong compression ratio of $256$, the latent space clearly captures the asymmetric von Kármán vortex street with distinct directionality, establishing a direct correspondence with the ground truth vorticity field. Compared to the latent field at $t^n$, the vortices are advected as expected at $t^{n+1}$; the change is subtle since only two consecutive vorticity snapshots are considered. Additionally, a clockwise vortex appears at the extreme downstream end in both latent flow fields, which is partially present in the ground truth. While this may initially appear spurious, it is indeed physically relevant, correctly reflecting the vorticity directionality (clockwise rotation). As anticipated, the decoder subsequently reconstructs the future vorticity field accurately using the marched latent field, as shown in Fig.~\ref{fig:temp_march_cylinder}c.

Next, the latent vorticity flow fields obtained from DIANO under different VTE formulation fidelities are discussed. Four simplified VTE formulations were considered for latent temporal evolution: 2D Stokes flow, i.e., linearized VTE without the convection term (Eq.~\ref{vte2}); 2D inviscid linearized VTE (Eq.~\ref{vte3}); 1D linearized VTE along the streamwise direction $x$ (Eq.~\ref{vte4}); and 1D linearized VTE along the normal direction $y$ (Eq.~\ref{vte5}), as shown in Figs.~\ref{fig:temp_march_cylinder}d--g. The spatial organization of latent vortical structures and reconstruction quality exhibit strong, simultaneous sensitivity to the physical fidelity of the embedded VTE formulation, depending on its degree of alignment with the underlying flow physics.  When the chosen VTE formulation is aligned with the advection-dominated, streamwise-convection character of the ground-truth dynamics ---specifically the 2D inviscid VTE and 1D VTE-$x$ (Eqs.~\ref{vte3} and~\ref{vte4}), yield latent space that preserves coherent vortical structures closely reproducing the von Kármán vortex street. Conversely, formulations that deviate from the underlying flow physics, such as 2D Stokes flow and 1D VTE-$y$, produce latent representations that lack physical coherence; discernible structures are present and visualizable, but would not correspond to physically meaningful flow features.  Quantitative reconstruction errors reported in Table~\ref{tab:hyper1} corroborate these trends, with errors ordered as: 2D linearized VTE $<$ 2D inviscid VTE $<$ 1D VTE-$x$ $<$ 2D Stokes flow $<$ 1D VTE-$y$. This hierarchy confirms a consistent and intrinsic coupling: the fidelity of the embedded VTE formulation relative to the true governing physics jointly determines the organization of coarse-grained latent vortical structures and the resulting reconstruction accuracy.

Overall, these results establish a clear principle: when the embedded latent PDE formulation is aligned with the true underlying flow physics, the latent space yields physically coherent and well-organized vortical structures with superior reconstruction accuracy. For advection-dominated flows, such as the cylinder wake at $\mathrm{Re} = 100$, retaining the advection term, and particularly the streamwise convection component, produces coherent and explainable latent vorticity fields closely resembling the input von Kármán vortex street, whereas the progressive removal of these physically essential terms leads to a systematic degradation in both latent structure organization and reconstruction accuracy.

\subsubsection{Generalization Across Reynolds Numbers}
\begin{figure}
    \centering
    \includegraphics[width=0.8\textwidth]{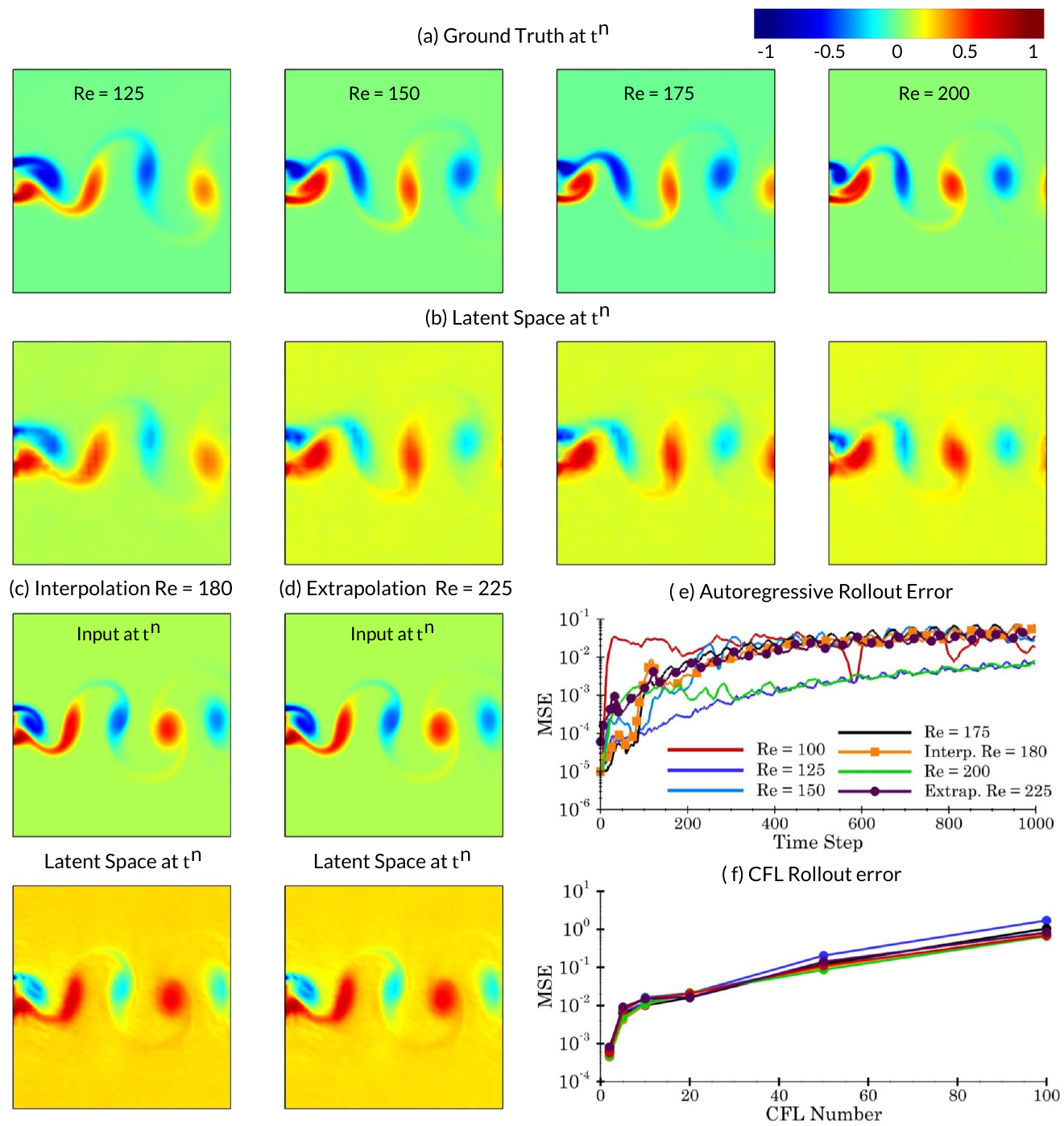}
    \caption{
    Generalization of the DIANO framework across Reynolds numbers.
    (a) Input vorticity fields at time $t^n$ from the test dataset corresponding to \textit{training} Reynolds numbers $\mathrm{Re} \in (125,150,175,200)$.
    (b) Corresponding latent vorticity fields at time $t^n$.
    (c) Input and latent vorticity fields at $t^n$ for \textit{interpolation} at $\mathrm{Re}=180$.
    (d) Input and latent vorticity fields at $t^n$ for \textit{extrapolation} at $\mathrm{Re}=225$.
    (e) Autoregressive rollout error over time across all Reynolds numbers.
    (f) Effect of increasing CFL number in the latent VTE solver.  The compression ratio and the Fourier modes are chosen to be 64 and 16, respectively.
    }
    \label{fig:temp_march_different_Re}
\end{figure}

In this section, we evaluate the ability of the DIANO framework to generalize across flows at varying Reynolds numbers. To this end, additional CFD simulations are performed for six Reynolds numbers, $\mathrm{Re} \in (125, 150, 175, 180, 200, 225)$, in addition to $\mathrm{Re} = 100$ considered in the preceding discussion. Among these, $\mathrm{Re} = 180$ and $\mathrm{Re} = 225$ are reserved for interpolation and extrapolation experiments, respectively, while the remaining cases are used for training. We focus on three key aspects: (i) the incorporation of Reynolds number variations in a physically consistent manner within the latent dynamics,  (ii) the ability of the DIANO framework to learn a parametric, explainable, and physically coherent latent vorticity representation across different Reynolds numbers, and (iii) quantitative robustness, evaluated through autoregressive rollout error (sequential one-step integration over the full time horizon) and performance under multiple CFL conditions (i.e., varying time-step sizes), across selected Reynolds numbers, as illustrated in Fig.~\ref{fig:temp_march_different_Re}.

In contrast to conventional ML approaches, where parametric dependence (such as the Reynolds number) is introduced through simple input augmentation, the DIANO framework allows this dependence to be embedded directly within the latent-space PDE solver (here, VTE, Eq.~\ref{vte1}). Rather than treating $\mathrm{Re}$ as an auxiliary feature, we enforce its influence through the viscosity parameter, while maintaining a fixed characteristic velocity scale, consistent with the relation $\nu \propto \mathrm{Re}^{-1}$. As a result, variations in Reynolds number are realized as intrinsic modifications to the latent governing equations, rather than as extrinsic inputs to a learned mapping. This enables the latent VTE solver to evolve flow fields under different dynamical $\mathrm{Re}$ regimes in a physically consistent manner. By enforcing parametric dependence at the level of the latent physics, DIANO may capture a continuous spectrum of flow behaviors, leading to improved interpolation within the training range and more reliable extrapolation beyond it, while avoiding the inconsistencies often associated with purely data-driven parameter conditioning.

By embedding the Reynolds number variation directly within the latent VTE solver, the DIANO framework enables explicit extraction and visualization of Reynolds-number-dependent latent vorticity fields across the considered range of \(\mathrm{Re}\) numbers. As illustrated in Fig.~\ref{fig:temp_march_different_Re}a, the input vorticity fields at time \(t^n\) for $\mathrm{Re} \in (125,\allowbreak 150,\allowbreak 175,\allowbreak 200)$ exhibit strong structural correspondence with their latent counterparts shown in Fig.~\ref{fig:temp_march_different_Re}b. This agreement indicates that the learned latent fields preserve essential flow features, including coherent structures such as the von K\'arm\'an vortex street, while maintaining a consistent correspondence and mapping between them. This consistency extends to both interpolated and extrapolated Reynolds numbers, as the input and the latent vorticity fields at \(t^n\) corresponding to \(\mathrm{Re} = 180\) and \(\mathrm{Re} = 225\), shown in Fig.~\ref{fig:temp_march_different_Re}c and Fig.~\ref{fig:temp_march_different_Re}d, respectively, remain coherent and physically meaningful, demonstrating that DIANO captures the underlying parametric dependence on \(\mathrm{Re}\) number in a continuous manner within the latent dynamics, thereby enabling reliable interpolation and moderate extrapolation. Nevertheless, this generalization is inherently $\mathrm{Re}$-number-regime-dependent. Since both the training and extrapolated Reynolds numbers lie well within the ``laminar periodic vortex-shedding regime'', the DIANO model effectively learns a regime-consistent representation of the flow dynamics. Consequently, for Reynolds numbers outside this regime (e.g., \(\mathrm{Re}\sim 1000\), where the flow transitions to turbulence), it is not expected that the model remains accurate.

Focusing on the quantitative behavior, the temporal predictive capability of the DIANO model is evaluated in two complementary analyses across all seven selected Reynolds numbers. First, an autoregressive rollout is performed. Starting from the state at $t^n$, the model advances one time step within the latent space to predict the solution at $t^{n+1}$, and the prediction is recursively used as input for subsequent steps. The resulting error evolution, shown in Fig.~\ref{fig:temp_march_different_Re}e, indicates that the rollout error initially grows and then saturates, leading to stable predictions at long times; however, the saturated error remains on the order of $O(10^{-2})$. Here, two distinct error growth behaviors are observed. For $\mathrm{Re} = 100$, the error increases rapidly, reaching $O(10^{-2})$ at earlier times, after which it saturates. In contrast, for the remaining Reynolds numbers, the error grows more gradually before approaching a similar saturated level. Furthermore, while both interpolated and extrapolated Reynolds number cases approach comparable error values at large times, the extrapolated cases exhibit higher errors during the initial transient evolution. A more detailed discussion of these error dynamics is provided in ~\ref{append2}. Second, the sensitivity to the latent time-integration step size is investigated by systematically varying the time step size in the latent VTE solver for a fixed input solution at $t^n$, thereby increasing the effective CFL number. As shown in Fig.~\ref{fig:temp_march_different_Re}f, the prediction error increases monotonically (log-scale) with the CFL number for all Reynolds numbers.

\subsubsection{Methodological Comparison}
\begin{figure}
    \centering
    \includegraphics[width=0.8\textwidth]{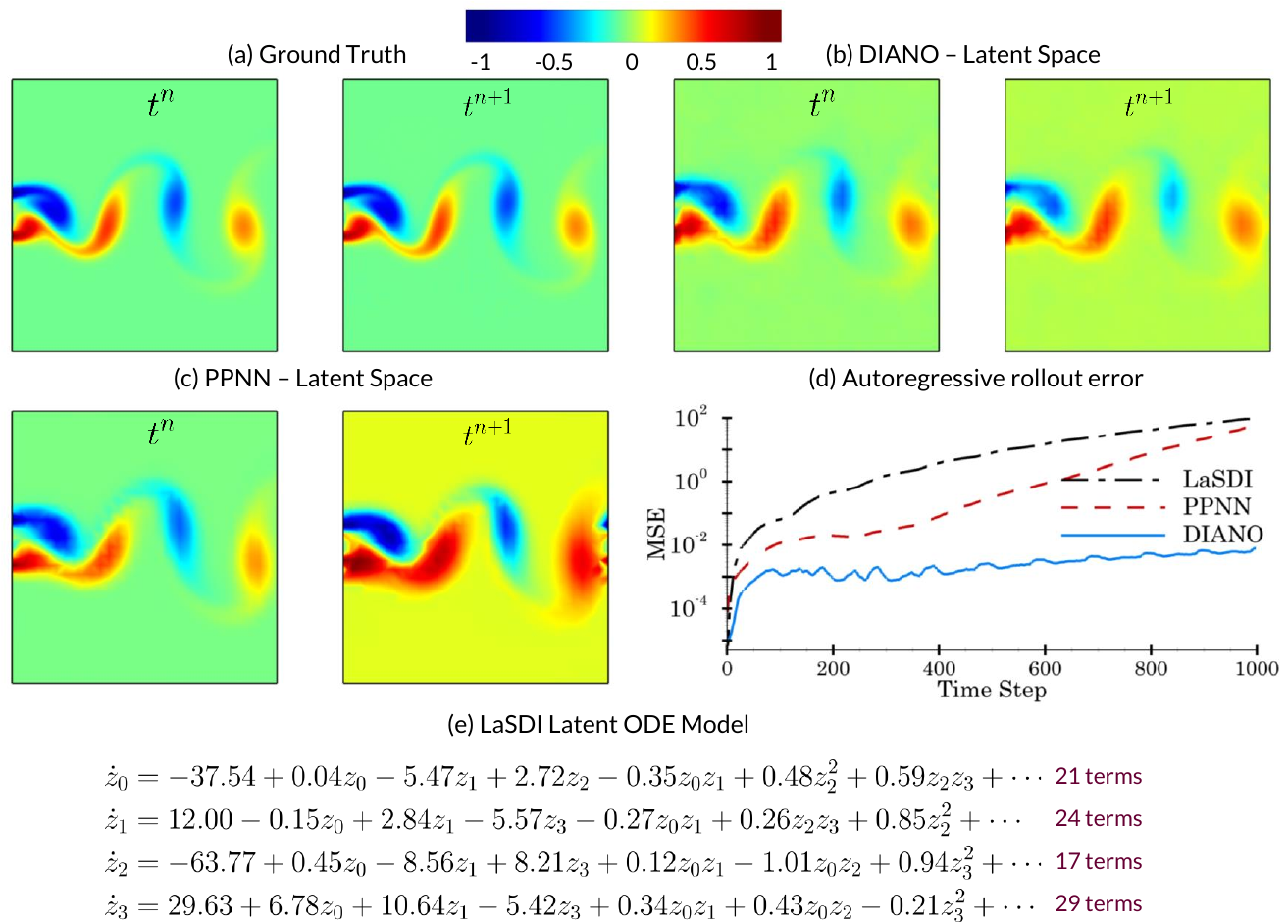}
    \caption{
    Methodological comparison across three frameworks: DIANO, PPNN, and LaSDI.
    (a) Ground-truth vorticity fields at times $t^n$ and $t^{n+1}$ from the \textit{test} dataset.
     Corresponding latent vorticity fields at times $t^n$ and $t^{n+1}$ of (b) DIANO and (c) PPNN methodologies.
    (d) Autoregressive rollout error comparison across the three methods.
    (e) Latent dynamical system identified for LaSDI using SINDy.}
    \label{fig:temp_march_methods}
\end{figure}

In this section, a methodological comparison is presented for the temporal marching benchmark considered in this work. Three approaches are evaluated: the proposed DIANO model, the Physics-Preserving Neural Network (PPNN)~\cite{liu2024multi}, and the LaSDI framework~\cite{fries2022lasdi}. The comparison investigates both the qualitative structure of the latent space and quantitative performance measured via autoregressive rollout error dynamics, as depicted in Fig.~\ref{fig:temp_march_methods}.

To ensure a consistent basis for comparison with the DIANO framework, which utilizes the 2D linearized VTE (Eq.~\ref{vte1}), the same governing equation, compression ratio (64), viscosity, velocity scale (corresponding to $\mathrm{Re}=125$), and time step for integration are used in the PPNN framework. In addition, common training settings, including the training and testing datasets, the optimizer, and the step-decay learning rate schedule, are kept consistent across the PPNN and LaSDI implementations. The PPNN method employs an autoencoder in which bilinear downsampling is applied to the input solution at $t^n$ to obtain a latent representation, then applies CNN-based stencil derivative operations to approximate the governing PDE, followed by bicubic upsampling and combined with a correction network to produce the solution at $t^{n+1}$. In contrast, the LaSDI framework adopts a fundamentally different strategy: first, it uses a NN-AE to perform a static mapping (using four latent modes in this comparison), followed by offline identification of an ODE latent dynamical system via LASSO regression or the SINDy methodology. During rollout, DIANO and PPNN both advance the solution autoregressively, using the predicted state at $t^n$ as input for $t^{n+1}$. In the LaSDI framework, the initial solution is first encoded into latent variables, evolved forward in time using the identified ODE system to a desired time $t$, and then decoded back to the physical space.

From the perspective of latent space structure, both DIANO and PPNN enable direct visualization of the latent vorticity field in physical space at time instances $t^n$ and $t^{n+1}$, as shown in Fig.~\ref{fig:temp_march_methods}b and Fig.~\ref{fig:temp_march_methods}c, respectively, alongside the corresponding ground truth solution depicted in Fig.~\ref{fig:temp_march_methods}a.  As expected, PPNN yields latent fields that remain physically coherent and meaningful, accurately capturing the vortex directionality characteristic of the von Kármán vortex street, with minor anomalies at the right boundary of the latent field at $t^{n+1}$, attributable to the use of \textit{periodic} derivative stencils. A key distinction, however, lies in how these representations are constructed. In PPNN, the latent fields are obtained through bilinear/bicubic resampling of the high-fidelity solution, whereas DIANO learns comparable coarse representations directly through its encoder operator. Despite this difference, both approaches preserve coherent vortical structures in the latent space. In contrast, the NN-AE-based LaSDI framework encodes the flow field into a set of scalar latent variables that lack direct spatial visualization. The discovered temporal latent dynamical system via the SINDy method is shown in Fig.~\ref{fig:temp_march_methods}e, consisting of polynomial terms up to second order and their corresponding active terms. A supplementary experiment further explored parameterized nonlinear dynamical systems with Fourier and exponential candidate functions, treating their frequencies and exponents as trainable parameters. Using the differentiable ADAM-SINDy framework~\cite{adam_sindy}, a more compact representation was obtained, reducing the number of active terms by approximately $14\%$, although these results are not shown here.

Quantitatively, the three frameworks exhibit distinctly different behaviors in autoregressive rollout, as illustrated in Fig.~\ref{fig:temp_march_methods}d. In terms of inference cost over the rollout duration shown, LaSDI is the most computationally efficient, requiring $2.279$ sec, compared to $8.046$ sec for PPNN, and $10.731$ sec for DIANO, measured on an NVIDIA RTX A4500 GPU. These differences reflect the increasing architectural complexity of each method: LaSDI leverages a low-dimensional ODE model for time advancement, whereas both PPNN and DIANO solve the more computationally expensive 2D linearized VTE in latent space. While the wall-clock costs of PPNN and DIANO are broadly comparable, DIANO incurs additional overhead due to its more sophisticated numerical treatment, wherein the latent PDE solver enables direct selection of discretization schemes rather than relying on CNN-based stencil approximations. In the present work, high-order implicit upwind compact schemes are employed for the advective terms, central differences for diffusion, and the RK4 method for time integration, thereby increasing the cost of differentiable PDE solutions compared to the forward Euler scheme adopted in PPNN. These architectural differences are further reflected in the long-term rollout behavior. Since LaSDI represents the spatial vorticity modes using low-dimensional latent scalar variables that do not explicitly embed the governing PDE, its rollout relies entirely on the identified dynamical system. As a result, the prediction error grows rapidly at early times and subsequently saturates, approaching a magnitude of $O(10^{2})$. Notably, it remains unaffected by misspecification of the latent governing equations, in contrast to DIANO, where the fidelity of the embedded PDE directly governs prediction quality. For the comparison setting, both PPNN and DIANO demonstrate improved stability during rollout; however, DIANO consistently maintains lower error levels than PPNN across all time steps, and although its error gradually increases over time, it saturates at the order of $O(10^{-2})$, indicating stable, albeit not perfectly accurate, long-term predictions. This improved behavior can be attributed to key elements of the DIANO framework. First, the integrated latent differentiable physics solver enables the use of physically consistent, high-accuracy numerical schemes, which improves stability during temporal integration.  Second, the DIANO architecture leverages an autoencoding neural operator, enabling mesh-invariant mappings between resolutions while learning compact coarse representations in the encoder and reconstructing high-fidelity fields in the decoder; it is hypothesized that, in this process, the \textit{explicit} spectral inductive bias is being introduced inherently by the adopted Fourier layers that preferentially retains essential vortical information through resolving dominant Fourier modes while attenuating the higher-frequency modes (small scale errors) successively, which may contribute to improved long-term stability and coherent propagation of flow structures during temporal rollout. 

%A few supplementary experiments were performed to investigate the robustness of the proposed DIANO framework. First, the fully coupled 2D nonlinear VTE (Eq.~\ref{vte}) was employed instead of the simplified VTE model. In this case, the DIANO framework failed to produce an interpretable latent space and showed poor reconstruction of the future solution. These results are therefore excluded, highlighting that the encoder effectively filters the essential small-scale flow structures necessary for stable and accurate numerical PDE evolution. 
%By leveraging the differentiability of the developed PDE solver, we attempted to learn a 2D linear VTE model governed in the latent space by making the viscosity and velocity scales trainable, rather than fixed (an inverse problem in the latent space). The results, provided in~\ref{append2}, demonstrate that the DIANO framework can capture the von Kármán vortex street in the latent space when both parameters are trained jointly. However, latent space interpretability was lost when either parameter was held fixed. It is important to emphasize that this experiment, focused on latent-space PDE discovery, primarily serves to \textit{illustrate} DIANO’s capability to identify governing equations during training, which lies beyond the scope of the present work. Investigating the limitations of latent interpretability when one parameter is fixed is an interesting avenue for future research.

\subsection{Geometrical Reduction with Temporal Marching}
\begin{figure}  
    \centering
    \includegraphics[width=0.8\textwidth,height=\textheight,keepaspectratio]{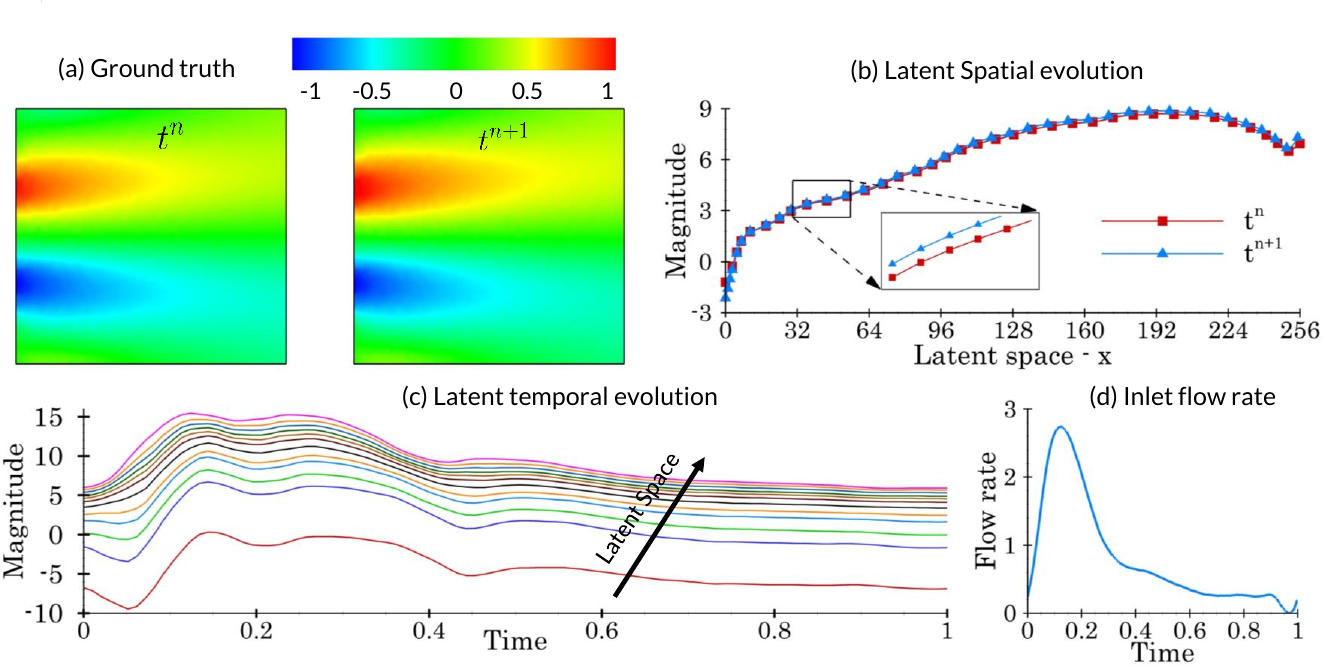}
    \caption{Geometrical Reduction with Temporal Marching. The 2D stenosis example is considered and spatio-temporal evolution of the vorticity latent space within the DIANO framework, computed using the 1D linearized VTE-$x$ formulation solved in the latent space. The transient flowfield from a \textit{test} dataset, containing two counter-rotating vortices downstream of the symmetric stenosis, is used for evaluation. 
    (a) Ground-truth vorticity fields at \( t^n \) and \( t^{n+1} \) from the test dataset. 
    (b) Corresponding 1D streamwise latent space representation advanced from \( t^n \) to \( t^{n+1} \) using the 1D linearized unsteady VTE - x (no normal derivative components). 
    (c) Temporal evolution of randomly selected latent features. 
    (d) Inlet flow-rate profile used for data generation.}
    \label{fig:symm}
\end{figure}

In this section, the geometrical reduction capability in conjunction with temporal marching is illustrated using our DIANO framework. In this approach, only the spatial dimensionality of the input function is reduced, while the temporal marching remains unaltered. Specifically, the encoder compresses the input at $t^n$ from geometric dimension $m$ to $m-1$, the latent representation is advanced in time to $t^{n+1}$ using the PDE solver, and the decoder restores the latent representation from $m-1$ back to $m$ dimensions, reconstructing the output at $t^{n+1}$. For a given two-dimensional vorticity field, two distinct strategies of geometrical reduction are explored: (i) compression along the normal direction (y), which produces a 1D latent vorticity distribution along the streamwise direction (x), and (ii) compression along the streamwise direction (x), which yields a 1D latent distribution along the normal direction (y). For temporal marching, the first scenario employs the differentiable PDE with 1D VTE along x (Eq.~\ref{vte4}), while the second uses 1D VTE along y (Eq.~\ref{vte5}), consistent with the respective spatial reduction directions. In this test case, we consider the 2D stenosed artery example (Fig.~\ref{fig:geometry}b). As in the previous objective, the differentiable VTE solver uses a characteristic velocity scale $V = 1$ and kinematic viscosity $\nu = 0.01$, chosen based on the Reynolds number, with a time step of $0.01$ for time integration. By explicitly separating the spatial compression directions, this approach highlights how low-dimensional spatial representations can efficiently represent the complex flow dynamics while preserving temporal evolution.

Figure~\ref{fig:symm} presents the results of the first spatial reduction scenario (compression along the normal direction) obtained using the DIANO framework, which incorporates a single Fourier layer with 64 Fourier modes in both the encoder and decoder. These results are based on ground truth flow fields from the test dataset, shown in Fig.~\ref{fig:symm}a at time instants \(t^n\) and \(t^{n+1}\), capturing the formation of two counter-rotating vortices downstream of the symmetric stenosis. The corresponding 1D streamwise latent vorticity representation at both time instants \(t^n\) and \(t^{n+1}\) is shown in Fig.~\ref{fig:symm}b. This latent representation is largely independent of the ground truth vortices, which is expected since the reduction is performed along the normal direction. Ideally, the latent vorticity along the centerline—the line dividing the two vortices—should exhibit a nearly constant value along the streamwise direction \(x\). However, the results show an appreciable slope in the upstream region (\(x < 32\)), while the downstream latent vorticity remains approximately constant, exhibiting only a subtle gradient along the streamwise direction.

The temporal latent evolution at four randomly selected streamwise points is shown in Fig.~\ref{fig:symm}c, together with the transient inlet flow profile in Fig.~\ref{fig:symm}d. In contrast to the spatial reduction, the temporal latent vorticity evolution closely follows the inlet flow profile,  reflecting the periodic forcing imposed by the boundary condition. The latent representation faithfully mimics the amplitude of these periodic flow variations, with only subtle phase shifts over the cardiac cycle. Collectively, these results demonstrate that temporal latent dynamics preserve vortex strength evolution in alignment with the inlet forcing, whereas spatial reduction captures only averaged streamwise features. For the second spatial reduction scenario—compression along the streamwise direction—the DIANO reconstructions of both spatial and temporal latent vorticity are qualitatively similar to those obtained under normal-direction compression, showing no additional distinctive features. Thus, a detailed visualization is omitted for brevity. Quantitatively, reconstruction accuracy is higher for the streamwise reduction case (Table~\ref{tab:hyper1}),  indicating that the direction of spatial reduction has a measurable influence on output reconstruction accuracy.

%In both geometrical reduction scenarios, the interpretability of spatial latents is limited, as vorticity information is strongly compressed relative to the corresponding temporal latents. This limitation may result from the symmetric geometry of the chosen stenosis. An asymmetric stenosis could produce characteristic flow asymmetries, such as a dominant anticlockwise vortex downstream, potentially enabling latent spatial encodings to capture more informative and physically meaningful flow features. Examining such configurations would help assess the capacity of spatial latent representations to encode relevant flow structures. 

\subsection{Many-to-One Functional Mapping}
\begin{figure}  
    \centering
    \includegraphics[width=0.8\textwidth,height=\textheight,keepaspectratio]{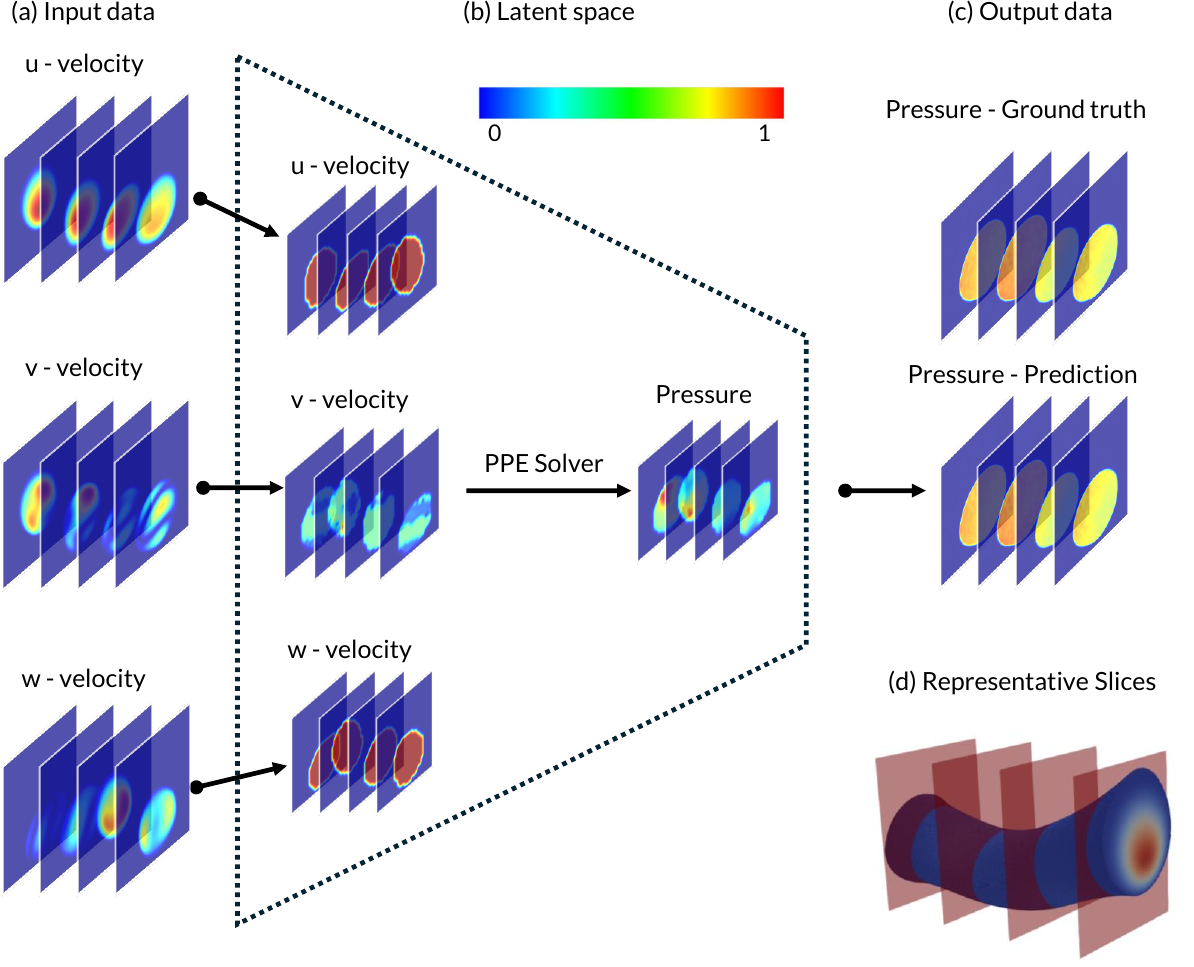}
    \caption{Many-to-One Functional Mapping. The DIANO framework infers a single pressure field from three input functions—three velocity components—using a differentiable PPE latent solver at the same time instant $t^n$. Flow through a patient-specific stenosed coronary artery is used for evaluation. (a) Ground truth velocity components—\(u\), \(v\), and \(w\)—from the \textit{test} dataset are independently encoded into latent space via three separate encoders, producing corresponding latent representations. (b) The 3D iterative differentiable PPE solver is applied within the latent space to obtain latent pressure, which is then reconstructed into high-dimensional 3D pressure data (c). (d) Representative 2D slices from the 3D grid are shown, following the flow direction from left to right.}
    \label{fig:lad}
\end{figure}

This section demonstrates the many-to-one function mapping capability of the DIANO framework, where the number of input functions exceeds the number of output functions. The demonstration focuses on computing a scalar pressure field from three velocity components using the Pressure Poisson Equation (PPE)-Eq.~\ref{ppe}, solved with an iterative elliptic PDE solver. Using the DIANO framework (8 Fourier modes and a compression ratio of 4), the three 3D velocity fields are independently compressed through three separate encoders, generating three corresponding 3D latent velocity representations. A differentiable PPE solver, implemented via the Point-Jacobi iterative method (tolerance $1\times10^{-6}$, maximum 150 iterations, and density $\rho$  of 1.06), computes the 3D latent pressure representation from these latent velocity representations. Subsequently, a single decoder reconstructs the output 3D pressure field at full resolution. This approach is similar to the DIANO framework used for nonlinear dimensionality reduction in static mappings. The key distinction lies in the many-to-one mapping where three input functions are mapped to a single output function via the PDE solver in the latent space.

In this experiment, the 3D flow field around a stenosed patient-specific coronary artery is considered. When mapping the original 3D unstructured data onto a structured grid suitable for the DIANO framework, ghost grid points with zero values were introduced outside the artery walls. The results of the DIANO framework are presented in Fig.~\ref{fig:lad}, which illustrates the three input velocity components and their corresponding latent representations, their mapping to the latent pressure field through the differentiable PPE solver, and the reconstruction of the output pressure field compared with the ground-truth pressure distribution.

Compared with the parabolic VTE solver, two limitations arise in comprehending the latent-space coarse-grained features in the present case. First, the iterative nature of the PPE solver where isotropic diffusion of pressure information is coupled with the upwinding of velocity information leads to weaker correspondence between less significant latent flow structures and the input velocity field. Second, masking is required during PPE iterations in the latent space to incorporate artery wall boundary information. Without this masking, the latent representation becomes contaminated by ghost regions, resulting in a lack of explainable flow structures in both the velocity and pressure latents. Despite these limitations, the latent pressure representation effectively highlights high-pressure regions but lacks in showing the expected pressure drop along the flow direction, likely due to mapping between two different functions, unlike temporal-marching scenarios, where the same function is mapped at successive time steps.

To evaluate the impact of the iterative nature of the PPE solver on velocity latent representations, an additional experiment was conducted by removing the Laplacian operator in Eq.~\ref{ppe}. In this configuration, the solver no longer performs iterative updates; it computes only the right-hand side (RHS) corresponding to velocity gradient terms, requiring the decoder to infer the output pressure field solely from RHS information. The resulting latent pressure representation retained a qualitatively similar structure (results are not shown) compared to the full-PPE case, indicating that the framework can capture key pressure features from velocity gradients alone. The reconstruction error relative to the full-PPE experiment was slightly worse in this case (Table~\ref{tab:hyper1}). Overall, these experiments demonstrate that, despite challenges from iterative solver dynamics and ghost points, the DIANO framework effectively performs many-to-one function mapping in the 3D example.

%This outcome highlights two critical limitations: the treatment of boundary points in latent space and the influence of ghost points in the ground-truth dataset.  

%The framework produces physically consistent latent pressure features and accurately reconstructs the full-resolution pressure field.
%motivating further exploration of strategies to improve latent space fidelity and interpretability in future work.

\section{Discussion}
\label{sec5}
High-dimensional spatiotemporal systems exhibit complex structures across multiple spatial and temporal scales. Capturing these dynamics in a data-driven framework requires latent representations that are both compact and physically meaningful. Conventional ML methods often prioritize predictive accuracy over physical fidelity, enforcing governing constraints only indirectly, which can produce latent spaces that are unconstrained by physics and difficult to interpret. The central challenge is thus to design latent representations that entail coarse-grained, compressed high-dimensional representations, while also preserving the dominant spatiotemporal dynamics, providing insight into the system’s governing structures. Our DIANO framework addresses these challenges by methodological integration of deterministic autoencoders, operator learning, and differentiable PDE solvers, resulting in a unified physics-constrained ML architecture.

The framework was evaluated across four representative scenarios to assess robustness and the interpretation of coarse-grained latent features. First, static mapped nonlinear dimensionality reduction demonstrates the compact encoding of high-dimensional flow fields while preserving spatial one-to-one correspondence. Second, temporal marching extends this capability by evolving latent representations via differentiable PDE solvers, enabling physically consistent predictions of future latent and full states. Third, geometrical reduction highlights its ability to maintain meaningful temporal dynamics even under geometrical space compression, allowing geometrically lower-dimensional PDEs to operate on the latent representations. Finally, many-to-one functional mapping illustrates the framework’s capacity to fuse multiple inputs into a coherent single output within the latent space. Collectively, these evaluations indicate that DIANO offers a promising framework for efficiently modeling complex spatiotemporal problems while supporting physics-embedded autoencoding operator learning.

%However, recent work on architectural extensions to neural operators~\cite{berner2025principled} remarks that operator learning architectures do \textit{implicitly} perform encoding–decoding transformations in their corresponding \textit{functional space}, rather than in the physical space. For instance, in FNOs, spectral convolutions act as encoders by truncating high-frequency Fourier modes, while subsequent linear transformations serve as decoders~\cite{berner2025principled}.

Integrating autoencoders with operator learning frameworks offers significant advantages by facilitating nonlinear dimensionality reduction while enabling mesh-invariant complex input–output function relationships~\cite{seidman2023variational}. By incorporating basis functions such as Fourier, Laplacian, or Wavelet functions within the autoencoder, high-dimensional input functions are first projected into a compact, coarse-grid latent space and then mapped to learn complex output functions, guided by either probabilistic or regularized latent objective formulations~\cite{bunker2025autoencoders}.  Similarly, our DIANO framework employs Fourier functions as a basis layer and performs autoencoding directly in the physical space (via upsampling and downsampling), importantly, without relying on any latent regularization objectives beyond the differentiable physics. This encoding-decoding methodology yields physically meaningful latent embeddings of input functions on a coarse grid. As illustrated in Fig.~\ref{fig:dim_red_cylinder}, this approach learns physically meaningful and coherent vortical structures in the latent space, preserving vortical features qualitatively within it and enabling more structured, one-to-one correspondences and explainable representations, while quantitatively the vorticity energy spectra computed from the latent vorticity fields closely agree with those of the ground-truth fields, thereby confirming the physical fidelity of the learned latent representations.    

Modeling temporal evolution in high-dimensional spatio-temporal systems introduces additional challenges, as stable and physically consistent predictions require more than compact latent embeddings. A common approach is to use ML architectures in the latent space, such as recurrent-based or neural ODEs, which parameterize discrete and continuous dynamics, respectively. Although effective in capturing temporal dependencies, these approaches provide no guarantee of reflecting the underlying physics unless it is constrained~\cite {dashtbayaz2025physics}, and the discovered latent ROMs, formulated as a system of ODEs, describe temporal evolution but lack spatial visualization and explanation. To overcome these limitations, DIANO adopts a decoupled learning strategy, in which temporal evolution is modeled through the novel integration of differentiable PDE solvers directly within the latent space, ensuring that temporal evolution adheres to governing physical laws and introducing little to no additional trainable parameters by alleviating the need for auxiliary ML architectures. This coupling yields the physics-constrained latent dynamics, the advantage absent in black-box latent ML models. Importantly, DIANO exploits the coarse-grid-like latent structure and supports the use of inexpensive lower-fidelity PDE solvers in the latent space, offering a flexible alternative to high-fidelity PDEs that represent the full physics. 
As discussed in Sec .~\ref {sec3}, the fidelity of the embedded PDE model directly influences the physical relevance of the latent mapping, particularly in recovering coherent vortical structures. A latent PDE fidelity that closely approximates the underlying physics is therefore \textit{recommended}, as it promotes both an explainable latent representation and superior reconstruction accuracy, establishing a principled trade-off between solver fidelity and reconstruction accuracy. Moreover, the generalization capability of the DIANO framework is demonstrated across various Reynolds numbers through a novel input augmentation strategy that embeds parametric (Reynolds number) variability directly within the latent PDE solver, exhibiting robust long-term stability during autoregressive temporal rollout in both interpolation and moderate extrapolation regimes.

A supplementary experiment is conducted to evaluate the impact of noise contamination on both static mapping and temporal marching tasks for the DIANO framework. Additive zero-mean Gaussian noise is introduced into the input fields, while the model is trained to recover the corresponding noise-free outputs, thereby formulating the problem as a denoising task. Detailed results are presented in~\ref{append3}. As expected, the reconstruction accuracy of the DIANO model relative to the clean functional mapping degrades in the presence of noise. Nevertheless, the framework remains robust to realistic noise levels (up to 15\%), preserving overall relevant latent representations and achieving satisfactory reconstruction accuracy. In contrast, at higher noise levels, the physical coherency of latent flow features deteriorated significantly, and the reconstruction error, as expected, increased; however, due to the denoising effect of the decoder, the errors remained acceptable.

Geometrical scaling also poses an open problem when applying ML models to geometrically large complex problems. The Universal Physical Transformer~\cite{alkin2024universal} frames this challenge as a \textit{problem-scaling} approach, mapping arbitrarily large input geometries to a fixed-size latent space, though interpretation is limited due to attention-based approximators at the bottleneck. A similar principle underlies physics-based PDE model reduction. For example, 1D blood flow models capture the cross-sectionally averaged behavior of full 3D PDE simulations, derived by averaging the Navier-Stokes equations across the vessel cross-section~\cite{csala2025physics}. Building on this concept, recent work on neural differential equations used 3D simulation data to enhance physics-based 1D blood flow models~\cite{csala2025physics}. Motivated by these ideas, we performed geometrical reduction experiments using the DIANO framework, which enables the discovery of novel geometrical reductions in a data-driven manner. Our results demonstrate that the dimensionality of input geometric data can be compressed by one order (e.g., 2D $\rightarrow$ 1D), enabling the solution of a lower-dimensional (1D) PDE defined on the reduced geometric space, while capturing the temporal coherence of flow dynamics. In particular, temporal latent evolution was driven by the inlet flow waveform, whereas spatial latents provided less physically meaningfulness, as expected due to geometric reduction. Our demonstration shows the DIANO framework simultaneously learns geometrical reductions from data while enforcing physics-based temporal evolution in the reduced geometric space.

In contrast to conventional operator learning tasks, which typically focus on one-to-one function mappings, relatively few approaches demonstrated many-to-one mappings, where multiple input fields map to a single output. Existing approaches include Multiple-Input FNO for elastodynamics, which concatenates multiple input fields sequentially~\cite{lehmann2025multiple}, DeepONet, which can combine multiple branch networks with a single trunk~\cite{jin2022mionet}, and meta-learning frameworks, which extend single-operator models to multi-operator tasks~\cite{sun2024lemon}. Although effective in establishing complex input-output relationships, these methods generally lack explicit physics constraints and explainable latent representations. In contrast, our experiments employ DIANO to perform many-to-one function mapping by inferring the pressure field from the three velocity components, governed by the 3D elliptic pressure-Poisson equation. DIANO accommodates multi-function inputs through independent encoders assigned to each velocity component; the resulting representations are subsequently fused and processed through a latent PDE model that produces a single latent pressure function. Notably, DIANO enables direct velocity-to-pressure mapping without explicitly solving the pressure Laplacian, which conventionally requires iterative numerical procedures. While the experiment is demonstrative, these results suggest that physics-guided operator learning offers a viable pathway for reducing the computational overhead associated with pressure-velocity coupling in incompressible flow solvers.

Future research on the DIANO framework can be pursued along several key directions. First, a fundamental limitation of the current framework is its reliance on inputs defined on well-structured grids, whereas ground-truth data are generated on unstructured grids. Future work should therefore aim to incorporate geometry-aware neural operator architectures into DIANO, apply PDE solvers to latent representations corresponding to unstructured grids, and evaluate how the inclusion of unstructured geometric information affects coarse-grained latent representations. Second, evaluating the framework’s performance in complex turbulent flows represents an interesting research direction. Unlike laminar flows considered herein, which exhibit a limited range of spatial scales, turbulent flows span a wide spectrum, from large geometrical features to fine-scale structures, such as those at the Kolmogorov scale. Investigating which spatial and temporal scales are preserved in the latent space could reveal DIANO’s multiscale capabilities and provide deeper insight into its performance in more complex flow scenarios. Third, the differentiability of the latent PDE solver opens a promising avenue for latent PDE discovery, where, rather than prescribing the governing equations a priori, treating the PDE parameters as trainable quantities enables the framework to directly identify the PDE characterizing the latent dynamics, transforming DIANO from a purely autoencoding operator into a data-driven tool for uncovering latent physics and motivating further investigation into equation discovery in compact latent spaces. Finally, integrating attention-based mechanisms within the autoencoding neural operators offers a particularly promising direction. Attention mechanisms could enable the framework to selectively focus on the most relevant spatial features, enhancing both predictive accuracy and scalability. Such enhancements may also improve capturing long-range dependencies in complex flows, supporting more effective multiscale representation learning and extending the DIANO framework’s applicability to challenging physical systems.

\section{Conclusion}
\label{sec6}
We introduced DIANO, a physics-integrated deterministic autoencoding operator learning framework that methodologically integrates autoencoders, neural operators, and differentiable PDE solvers for modeling high-dimensional spatiotemporal flows. Across benchmark tasks, DIANO achieves physically meaningful and explainable latent representations, accurate reconstructions, and physics-constrained temporal evolution, while efficiently compressing complex dynamics. Unlike conventional neural operators or purely data-driven models, DIANO combines PDE-constrained latent evolution with operator learning, producing coarse-grid, predictive, and mechanistically meaningful latent representations. This scalable and robust framework bridges data-driven operator learning and physics-based modeling, providing a new computationally efficient and accurate tool for analyzing, predicting, and understanding complex spatiotemporal systems across diverse scientific applications.

\section*{Conflict of Interest}
The authors declare no conflict of interest.

\section*{Acknowledgments}
This research was supported by the National Science Foundation under Award No. 2247173.

\section*{Data Availability} 
The DIANO framework, developed in the PyTorch environment, is publicly available on GitHub at \url{https://github.com/siva-viknesh/DIfferentiable_Autoencoding_Neural_Operator}.
\appendix
\section{Autoencoding architecture}\label{append1}
\subsection*{Fully Connected Neural Network Autoencoder}
The fully connected neural network autoencoder (NN-AE) is employed as a baseline for nonlinear dimensionality reduction of vectorized flow snapshots. The model adopts a symmetric encoder–decoder structure: the encoder compresses high-dimensional inputs into a compact latent space through successive fully connected layers of decreasing width, while the decoder mirrors this process to reconstruct the original input dimensionality. Nonlinear activation functions of ReLU are applied between layers. Although lacking spatial inductive biases, the NN-AE provides a flexible framework for learning compact latent embeddings and has been widely applied in developing ROMs of fluid systems~\cite{murata2020nonlinear}.

The NN-AE architecture follows the implementation in~\cite{csala2022comparing}. Both encoder and decoder are symmetric, with layer widths decreasing in the encoder and increasing in the decoder. Specifically:  

\[
\text{Encoder: } x_i \;\rightarrow\; 2048 \;\rightarrow\; 512 \;\rightarrow\; 128 \;\rightarrow\; x_l \;,
\]  
\[
\text{Decoder: } x_l \;\rightarrow\; 128 \;\rightarrow\; 512 \;\rightarrow\; 2048 \;\rightarrow\; x_i \;,
\]  
where $x_i$ is the input dimensionality (flattened snapshot size) and $x_l$ is the latent dimension. For the present work, the latent dimension $x_l$ is varied across $\{8, 16, 32\}$ to assess the effect of different compression ratios on reconstruction performance. Prior to training, snapshots are flattened into one-dimensional vectors and also normalized.

\subsection*{Convolutional Neural Network Autoencoder}
The convolutional neural network autoencoder (CNN-AE), originally developed for image processing tasks, has been used for nonlinear dimensionality reduction of flow dynamics, particularly in cases involving structured grid data. Unlike the fully connected NN-AE, which treats inputs as uncorrelated vectors, the CNN-AE explicitly leverages local spatial correlations through convolutional operations, making it well-suited for field variables~\cite{morimoto2021convolutional}.  

A similar CNN-AE architecture used in~\cite{fukami2020convolutional}, is employed in this study. The encoder maps the high-dimensional input flow field into a low-dimensional latent representation using six convolutional layers. The first three layers employ stride-2 convolutions with $3 \times 3$ kernels and padding of 1, progressively reducing the spatial resolution while increasing the channel depth from 1 $\rightarrow$ 64 $\rightarrow$ 128 $\rightarrow$ 256. The subsequent three layers utilize stride-1 convolutions to decrease the channel depth from 256 $\rightarrow$ 128 $\rightarrow$ 64 $\rightarrow$ 1, while preserving spatial resolution. The decoder is a mirrored version of the encoder, implemented with transposed convolutions. Its first three layers use stride-1 transposed convolutions to increase channel depth from 1 $\rightarrow$ 64 $\rightarrow$ 128 $\rightarrow$ 256, followed by three stride-2 transposed convolutional layers that reduce the channel depth from 256 $\rightarrow$ 128 $\rightarrow$ 64 $\rightarrow$ 1. Output padding is applied to restore the original spatial dimensions. The SiLU (Sigmoid Linear Unit) activation function is applied after all layers except the final output layer.

\textit{Residual connections:} To enhance feature extraction and improve the organization of latent representations, residual connections are incorporated into the CNN-AE architecture. Specifically, residual blocks are introduced after selected convolutional and transposed convolutional layers in both the encoder and decoder. Each residual block consists of two $3 \times 3$ convolutional layers with unit stride and padding, interleaved with SiLU activation functions, and includes an identity skip connection that adds the input feature map to the transformed output. This design facilitates improved gradient flow during training and enables the network to learn corrections to intermediate feature representations, rather than entirely new mappings.

In the encoder, residual blocks are applied at multiple resolution levels following convolutional layers, allowing progressive refinement of spatial features as the representation is compressed. Similarly, in the decoder, residual blocks are introduced after upsampling operations to recover coherent flow structures during reconstruction. This residual formulation, inspired by deep residual networks, improves the ability of the CNN-AE to capture coherent vortical patterns and yields more structured and physically meaningful latent representations compared to a purely feedforward architecture.

\begin{figure}  
    \centering
    \includegraphics[width=0.8\textwidth,height=\textheight,keepaspectratio]{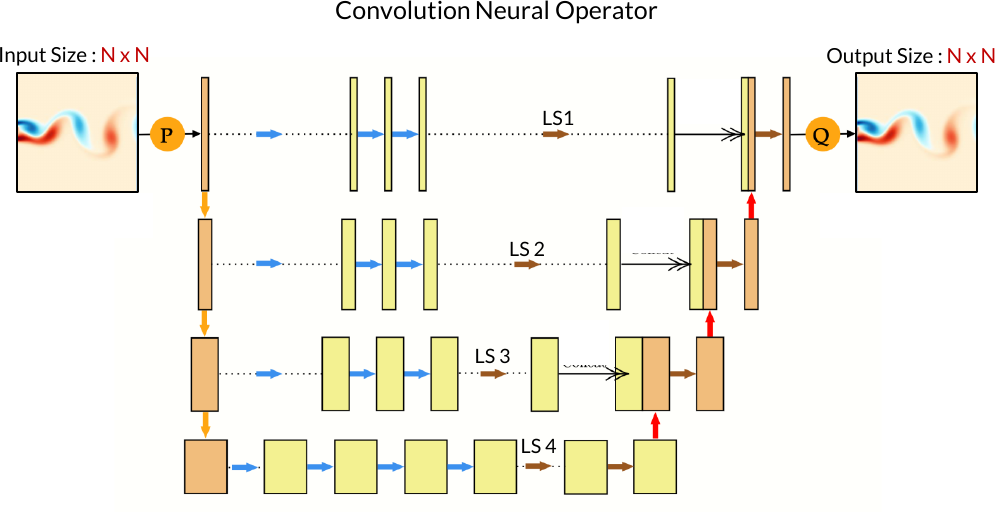}
    \caption{Convolutional Neural Operator (CNO) employs a U-shaped autoencoding strategy, producing four latent spaces (LS1–LS4) with progressively increasing compression. Each level reduces the feature dimensionality by a factor of 2.}

    \label{fig:cno}
\end{figure}

\subsection*{Convolutional Neural Operator}
Convolutional Neural Operator (CNO)~\cite{raonic2023convolutional} is a U-Net–based architecture designed to learn mappings between infinite-dimensional function spaces. It combines autoencoding-based dimensionality reduction with invariant and residual skip connections, yielding meaningful latent spaces at multiple compression ratios in a single forward pass. By also adhering to the operator-learning principle, the CNO generalizes across mesh resolutions while maintaining stability and accuracy. In this work, we adopt the original CNO implementation, detailed in Fig.~\ref{fig:cno}, which follows a hierarchical encoder–decoder design with residual connections for stable multi-scale operator learning.

\textit{Encoder:} The encoder hierarchically downsamples the input field through a series of convolutional modules. Each block consists of two successive $3 \times 3$ convolutional layers with padding of 1, each followed by batch normalization and ReLU activation. A $2 \times 2$ max-pooling layer reduces spatial resolution between stages. Feature channel increases as $64 \rightarrow 128 \rightarrow 256 \rightarrow 512 \rightarrow 1024$, generating progressively abstract feature maps. The final layer, with a single feature channel, forms the bottleneck representation, serving as the compact latent space.

\textit{Latent space:} The architecture naturally constructs multi-resolution latent spaces (LS) with compression factors of $2$, $4$, $8$, and $16$, corresponding to downsampling levels LS1--LS4 indicated in Fig.~\ref{fig:cno}. These embeddings allow the model to capture both local fine-scale structures and global dependencies, which are critical for operator learning in high-dimensional function spaces.  

\textit{Decoder:} The decoder reconstructs the output field using $2 \times 2$ transposed convolutions with stride 2 for upsampling. At each stage, the upsampled features are concatenated with the corresponding encoder outputs through skip connections, ensuring the preservation of fine-grained spatial information. The fused features are then processed by a convolutional block of the same form as in the encoder. The number of channels decreases symmetrically ($1024 \rightarrow 512 \rightarrow 256 \rightarrow 128 \rightarrow 64$), and the final prediction is produced by a $1 \times 1$ convolution layer.

\textit{Invariant and residual blocks:} Invariant blocks (two-layer convolutional modules at fixed resolution) enable deeper feature extraction without altering spatial size. Residual connections, invoked at each convolutional stage, stabilize training, improve gradient propagation, and preserve high-dimensional latent representations. Together, they enhance multi-scale representation learning and enable the model to capture complex nonlinear mappings.

It is important to note that, although the CNO employs a U-shaped autoencoding strategy, its latent spaces have access to relatively high-dimensional representations, which is achieved by incorporating residual blocks after each convolutional operation, effectively mimicking multigrid numerical solvers such as the V-cycle and W-cycle. These cycles efficiently resolve flow-field scales ranging from fine to coarse at different stages and spatial resolutions, thereby enhancing both accuracy and computational efficiency. However, the multiple latent spaces together with skip connections make interpretation more difficult. 

\section{Rollout Error Dynamics}\label{append2}

\begin{figure}
    \centering
    \includegraphics[width=0.75\linewidth]{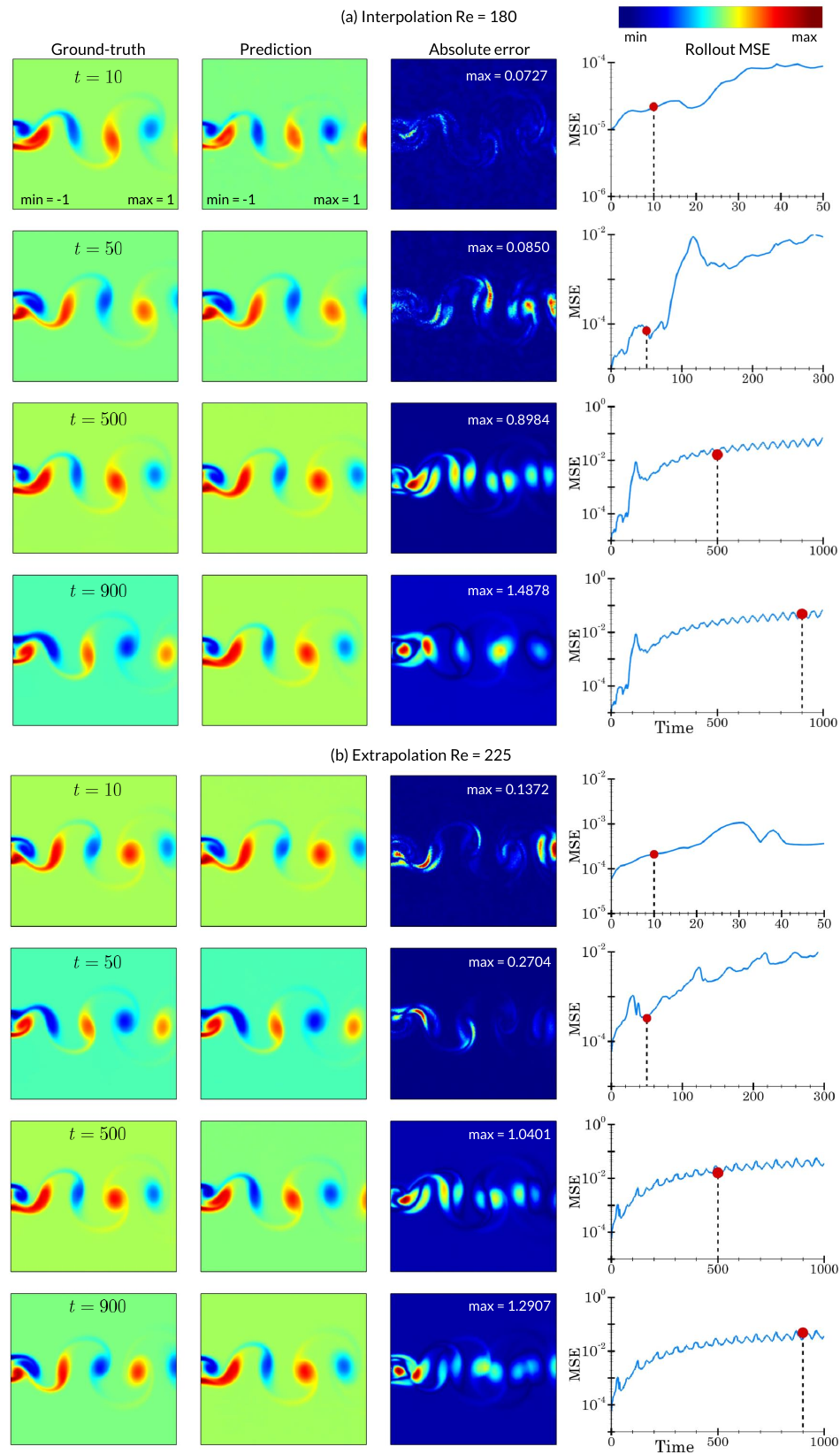}    
    \caption{Autoregressive rollout error dynamics for \textit{unseen} $\mathrm{Re}$ case: ground truth, predictions, pointwise absolute errors (min = 0), and rollout MSE at selected snapshots. (a) Interpolation $\mathrm{Re}=180$. (b) Extrapolation $\mathrm{Re}=225$.}
    \label{fig:error_Re_combined}
\end{figure}
\begin{figure}  
    \centering
    \includegraphics[width=0.8\textwidth,height=\textheight,keepaspectratio]{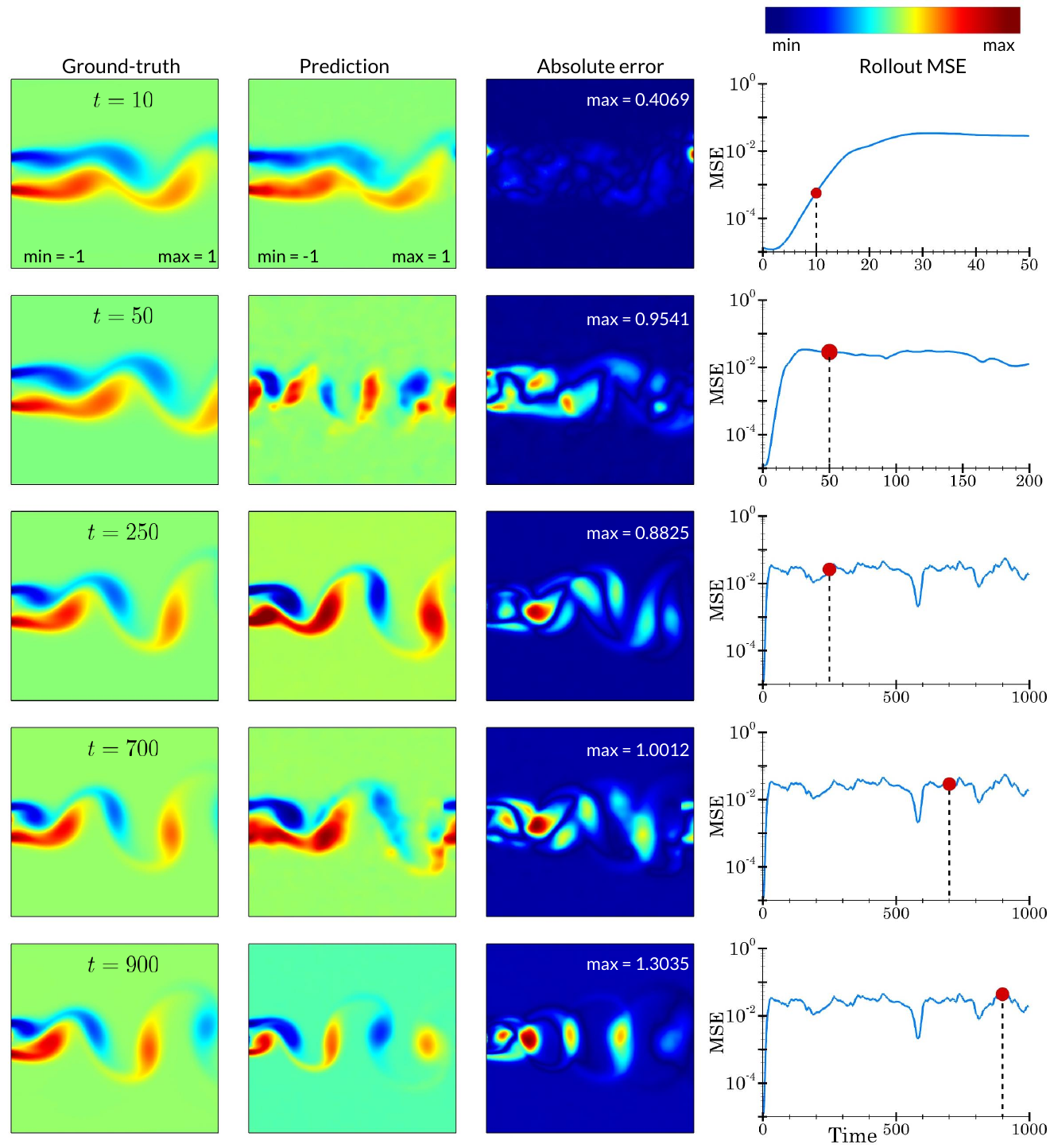}
    \caption{Autoregressive rollout error dynamics for $\mathrm{Re}=100$ (unbalanced case). Comparison between the ground-truth, predicted solutions, and the corresponding pointwise absolute error (min = 0), along with the temporal rollout MSE, with markers, indicating the chosen snapshot time index.}
    \label{fig:error_Re_100}
\end{figure}

In this section, we investigate two distinct error-growth behaviors exhibited by the DIANO framework during autoregressive temporal rollout across seven Reynolds numbers, as illustrated in Fig.~\ref{fig:temp_march_different_Re}e. Notably, the error accumulation at $\mathrm{Re}=100$ deviates significantly from the trends observed in the other higher six $\mathrm{Re}$ cases (including both interpolation and extrapolation), primarily due to transitional vortex dynamics in the training dataset, which result in an imbalance in temporal rollout predictions. To better understand the underlying error dynamics, we focus on two representative scenarios: (i) the unbalanced dataset corresponding to $\mathrm{Re}=100$, and (ii) a balanced dataset corresponding to unseen regimes, namely, interpolation at $\mathrm{Re}=180$ and extrapolation at $\mathrm{Re}=225$. Across all Reynolds numbers, the rollout MSE increases to $\mathcal{O}(10^{-2})$ and subsequently saturates, indicating bounded long-term prediction error, but with markedly different temporal evolution: at $\mathrm{Re}=100$, the error exhibits rapid initial amplification over the initial rollout steps before reaching the plateau, whereas for the remaining Reynolds numbers, including both interpolation and extrapolation, the error accumulates gradually without a pronounced transient phase, consistent with the gradual compounding of stepwise prediction discrepancies. Autoregressive rollouts for the unseen Reynolds numbers and $\mathrm{Re}=100$ are presented in Figs.~\ref{fig:error_Re_combined} and~\ref{fig:error_Re_100}, showing the ground-truth and predicted vorticity fields at time level $t^{n+1}$, the corresponding absolute error fields, and the MSE rollout curves with selected time markers.

\textit{Balanced Dataset:} For the interpolation case ($\mathrm{Re}=180$) and the extrapolation case
($\mathrm{Re}=225$), the flow remains in a fully developed, periodic vortex-shedding regime throughout the entire rollout window, as confirmed by the ground-truth vorticity fields in Figs.~\ref{fig:error_Re_combined}a and~\ref{fig:error_Re_combined}b. Despite sharing the same vortex-shedding flow regime, the two Reynold numbers exhibit quantitatively distinct error trajectories over time. At early rollout times ($t = 10$), both predictions are qualitatively faithful to the reference vorticity fields, with the predicted vortex street reproducing the correct spatial organization and vortex directionality. However, the extrapolation case already incurs an absolute error nearly twice as large (max $\approx 0.14$) as the interpolation case (max $\approx 0.07$), with discrepancies concentrated at the leading vortex positions and the immediate near-wake of the cylinder. This early-time gap indicates that the model struggles to precisely reproduce the shedding dynamics at the extrapolated $\mathrm{Re}$ case from the very first autoregressive step, even though the large-scale flow structure remains intact. As the rollout advances, the dominant error mechanism in both interpolation and extrapolation cases arises more from progressive phase shifts and positional drift of the shed vortices than from inaccuracies in their amplitude or core structure. Small stepwise discrepancies in vortex convection speed and shedding frequency accumulate across successive autoregressive steps, leading to the predicted vortex street gradually desynchronizing spatially with the reference field. By $t = 500$, this dephasing manifests as a train of error lobes aligned with the vortex positions, with maximum absolute errors of order unity. The dephasing proceeds at different rates: the rollout MSE saturates at $\mathcal{O}(10^{-2})$ earlier in the extrapolation case than in the interpolation case, consistent with faster accumulation of phase errors under out-of-distribution generalization.

\textit{Unbalanced Dataset:} For $\mathrm{Re}=100$, as shown in Fig.~\ref{fig:error_Re_100}, the MSE rises steeply within the first $t \approx 10$ autoregressive steps, reaching the $\mathcal{O}(10^{-2})$ saturation level almost immediately. Interestingly, the maximum absolute error at this early time ($\max \approx 0.41$) is substantially larger than that for both the interpolation ($\max \approx 0.07$) and extrapolation ($\max \approx 0.14$) cases, despite $\mathrm{Re} = 100$ lying within the training distribution. Inspection of the predicted and ground-truth vorticity fields clarifies this behavior: at $t = 10$, the ground-truth exhibits a smooth, elongated vortex street characteristic of developing low-Reynolds-number shedding, with well-separated alternating vortex cores, while the predicted field correctly captures the alternating directionality and spatial wavelength. However, by $t = 50$, positional mismatches have grown such that the predicted and reference vortex streets are nearly in antiphase—the ground-truth corresponds to the transitional/developing stage, whereas DIANO predicts fully developed vortices—producing pronounced error lobes across the domain with $\max \approx 0.95$.

At later times $t \in (250, 700, 900)$, the qualitative nature of the discrepancy remains similar: the predicted field continues to reproduce the correct fully developed vortex-shedding topology, while the absolute error remains concentrated in compact blobs aligned with the vortex cores, with maximum values ranging from $0.88$ to $1.30$. Notably, at $t = 700$, a structural deviation appears as the model predicts non-coherent vortex features while the ground-truth remains fully developed, likely reflecting  the residual memory of transitional dynamics within the model; by $t = 900$, the predictions better align with the correct periodic shedding, and as the flow fully enters the stable shedding regime, error growth becomes bounded, producing the observed MSE saturation. Importantly, this early-time error arises primarily from the presence of transitional vortex shedding states in the $\mathrm{Re}=100$ dataset, rather than from phase shifts or positional offsets of the vortex street, which dominate error accumulation at higher Reynolds numbers. In contrast, at higher $\mathrm{Re}$, the flow remains fully developed throughout the rollout, so the model operates entirely within a well-represented region of state space, and error growth is driven mainly by small phase shifts and spatial misalignments that compound over successive steps. Consequently, the early-time errors at $\mathrm{Re}=100$ could be mitigated by constructing the dataset using a temporal window that emphasizes fully developed vortex shedding, rather than applying a uniform time interval across all Reynolds numbers.

Across all three $\mathrm{Re}$ cases, the rollout MSE saturates at $\mathcal{O}(10^{-2})$ and remains bounded over the full temporal window, with no evidence of unbounded error growth. Each case exhibits a distinct error-growth mechanism: in the interpolation regime, errors accumulate gradually due to small stepwise discrepancies in convection speed and shedding frequency; in the extrapolation regime, errors grow more rapidly because of distributional shift and accelerated dephasing; and in the unbalanced regime, an immediate phase offset due to transitional dynamics during early rollout, after which predictions evolve on the correct flow topology but at a persistent spatial lag from the reference. Despite these differences, the large-scale flow structure, including vortex shedding topology, spatial wavelength, and convection speed, is preserved in all cases, indicating that phase drift, rather than structural breakdown, limits long-horizon prediction accuracy. We can attribute this bounded error growth to the explicit spectral inductive bias of the Fourier layers within the autoencoding architecture: by retaining only a finite set of low-wavenumber modes, these layers act as an implicit low-pass filter at each autoregressive step, suppressing the amplification of high-wavenumber errors that would otherwise destabilize long-term predictions. The consistent saturation of the MSE across seen, interpolated, and extrapolated Reynolds numbers suggests that this spectral truncation establishes an effective error saturation determined by the resolved wavenumber content of the architecture together with the integrated differentiable physics in the latent space.

\section{Robustness to Gaussian Noise}\label{append3}
In this section, we evaluate the robustness of the DIANO framework to measurement noise contamination, a scenario frequently encountered in real-world experimental data. To this end, we consider two spatio-temporal modeling tasks: (i)~\textit{static mapping}, which reconstructs the vorticity field at a fixed time instant $t^n$, and (ii)~\textit{temporal marching}, which advances the solution from $t^n$ to $t^{n+1}$. In both tasks, zero-mean additive Gaussian white noise with a prescribed standard deviation is superimposed on the input vorticity field, while the model is trained to recover the corresponding clean (noise-free) output, effectively framing the problem as a denoising task. For these experiments, we choose the Reynolds number $\mathrm{Re} = 125$ flow over cylinder case, employ a compression ratio of $\mathrm{CR} = 64$, and retain $16$ Fourier modes in the Fourier layers for both tasks, with the 2D linearized VTE used specifically for temporal marching. Under these settings, the framework is tested on input noise levels of $5\%$, $15\%$, and $25\%$, allowing us to investigate the influence of both moderate and severe perturbations.

\begin{figure}
    \centering
    \includegraphics[width=0.75\textwidth,height=\textheight,keepaspectratio]{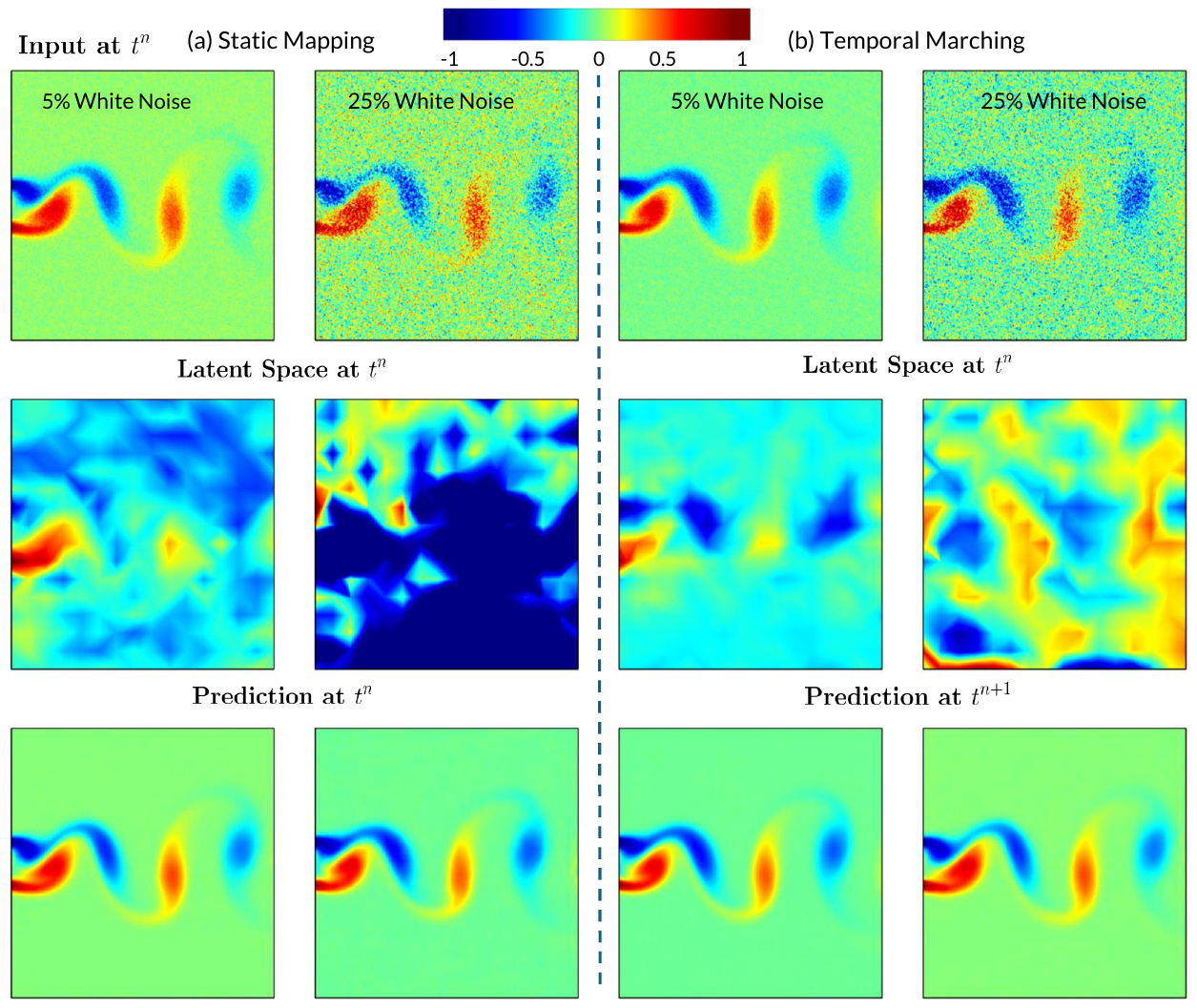}
    \caption{Gaussian noise experiment on reconstructing the clean von Kármán vortex street from the \textit{test dataset}. (a) Static mapping: showing the noisy vorticity field, the latent fields, and its reconstruction at the same instant $t^n$. (b) Temporal marching: showing the same noisy vorticity field and latent vorticity at $t^n$, and predicting the vorticity field at $t^{n+1}$. Results correspond to both 5\% (first panel) and 25\% (second panel) noise levels.}
    \label{fig:noise}
\end{figure}

\begin{table}
\centering
\begin{tabular}{|c|cc|cc|}
\hline
\multirow{2}{*}{Noise level} & \multicolumn{2}{c|}{Static Mapping} & \multicolumn{2}{c|}{Temporal Marching} \\ \cline{2-5} 
                             & \multicolumn{1}{c|}{Train Error} & Test Error & \multicolumn{1}{c|}{Train Error} & Test Error \\ \hline
5\%                          & \multicolumn{1}{c|}{$4.3194 \times 10^{-5}$} & $4.7123 \times 10^{-5}$ & \multicolumn{1}{c|}{$1.3986 \times 10^{-5}$} & $1.5019 \times 10^{-5}$ \\ \hline
15\%                         & \multicolumn{1}{c|}{$4.9604 \times 10^{-5}$} & $5.4086 \times 10^{-5}$ & \multicolumn{1}{c|}{$3.1676 \times 10^{-5}$} & $4.1839 \times 10^{-5}$ \\ \hline
25\%                         & \multicolumn{1}{c|}{$1.1838 \times 10^{-4}$} & $1.3048 \times 10^{-4}$ & \multicolumn{1}{c|}{$2.1616 \times 10^{-4}$} & $3.4759 \times 10^{-4}$ \\ \hline
\end{tabular}
\caption{Train and test MSE of the DIANO framework (compression ratio CR = 64, 16 Fourier modes) for static mapping and temporal marching using 2D linearized VTE  under additive zero-mean Gaussian noise, with the noise level denoting the standard deviation.}
\label{tab:noise_table}
\end{table}

Figure~\ref{fig:noise} presents representative results at both $5\%$ and $25\%$ noise levels, showing the noisy input vorticity fields alongside the corresponding latent representations and final predictions for static mapping (Fig.~\ref{fig:noise}a) and temporal marching (Fig.~\ref{fig:noise}b). In both tasks, DIANO successfully reconstructs a clean vorticity field from the corrupted input. Notably, despite the presence of noise, the latent vorticity fields at the lower noise level can retain the overall coherent von~Kármán vortex-shedding pattern observed in the ground-truth data, indicating that the encoder filters some of the stochastic perturbations during the compression step. The physical coherence of the latent structures is, of course, degraded compared to the noise-free baseline. As the noise level increases further, this degradation becomes more pronounced, and at high noise levels, the latent vorticity fields lose coherent vortex structures entirely, reflecting a breakdown in recovering the essential latent features under severe input perturbations. Interestingly, the predictions (decoded results) remain qualitatively accurate even at higher noise levels. 

Table~\ref{tab:noise_table} quantifies the impact of input noise on the predictive performance of the DIANO framework for both static mapping and temporal marching. At a moderate noise level of $5\%$, the framework achieves low training and test errors, with MSEs on the order of $10^{-5}$ for both tasks, reflecting its ability to accurately recover clean vorticity fields despite the presence of stochastic perturbations. As the noise level increases to $15\%$, the errors rise slightly, yet remain within the same order of magnitude, indicating that the framework maintains robust predictive accuracy. At $25\%$ noise, a more noticeable degradation occurs, particularly for temporal marching, where the test MSE increases to $3.48 \times 10^{-4}$. This trend aligns with the qualitative observations of the latent vorticity fields: at high noise levels, coherent vortex structures in the latent space become increasingly distorted and eventually lose their physical coherent organization, demonstrating that the cumulative effect of noise can propagate through the compression and temporal evolution steps. Collectively, these results highlight that while DIANO is robust to realistic noise levels, the preservation of organized latent vortical structure is progressively challenged as input noise becomes severe, emphasizing the trade-off between denoising performance and meaningful latent-space structures in noisy environments.

\bibliographystyle{elsarticle-num}
\bibliography{reference}

\begin{thebibliography}{10}
\expandafter\ifx\csname url\endcsname\relax
  \def\url#1{\texttt{#1}}\fi
\expandafter\ifx\csname urlprefix\endcsname\relax\def\urlprefix{URL }\fi
\expandafter\ifx\csname href\endcsname\relax
  \def\href#1#2{#2} \def\path#1{#1}\fi

\bibitem{lumley1967structure}
J.~L. Lumley, The structure of inhomogeneous turbulent flows, Atmospheric Turbulence and Radio Wave Propagation (1967) 166--178.

\bibitem{berkooz1993proper}
G.~Berkooz, P.~Holmes, J.~L. Lumley, The proper orthogonal decomposition in the analysis of turbulent flows, Annual Review of Fluid Mechanics 25~(1) (1993) 539--575.

\bibitem{schmid2010dynamic}
P.~J. Schmid, Dynamic mode decomposition of numerical and experimental data, Journal of Fluid Mechanics 656 (2010) 5--28.

\bibitem{zhang2015machine}
Z.~J. Zhang, K.~Duraisamy, Machine learning methods for data-driven turbulence modeling, in: 22nd AIAA Computational Fluid Dynamics Conference, 2015, p. 2460.

\bibitem{ling2016reynolds}
J.~Ling, A.~Kurzawski, J.~Templeton, Reynolds averaged turbulence modelling using deep neural networks with embedded invariance, Journal of Fluid Mechanics 807 (2016) 155--166.

\bibitem{kutz2017deep}
J.~N. Kutz, Deep learning in fluid dynamics, Journal of Fluid Mechanics 814 (2017) 1--4.

\bibitem{eivazi2020deep}
H.~Eivazi, H.~Veisi, M.~H. Naderi, V.~Esfahanian, Deep neural networks for nonlinear model order reduction of unsteady flows, Physics of Fluids 32~(10) (2020).

\bibitem{roweis2000nonlinear}
S.~T. Roweis, L.~K. Saul, Nonlinear dimensionality reduction by locally linear embedding, science 290~(5500) (2000) 2323--2326.

\bibitem{csala2022comparing}
H.~Csala, S.~Dawson, A.~Arzani, Comparing different nonlinear dimensionality reduction techniques for data-driven unsteady fluid flow modeling, Physics of Fluids 34~(11) (2022).

\bibitem{li2024manifold}
R.~Li, S.~Song, Manifold learning-based reduced-order model for full speed flow field, Physics of Fluids 36~(8) (2024).

\bibitem{erichson2020shallow}
N.~B. Erichson, L.~Mathelin, Z.~Yao, S.~L. Brunton, M.~W. Mahoney, J.~N. Kutz, Shallow neural networks for fluid flow reconstruction with limited sensors, Proceedings of the Royal Society A 476~(2238) (2020) 20200097.

\bibitem{agostini2020exploration}
L.~Agostini, Exploration and prediction of fluid dynamical systems using auto-encoder technology, Physics of Fluids 32~(6) (2020).

\bibitem{fukami2020convolutional}
K.~Fukami, T.~Nakamura, K.~Fukagata, Convolutional neural network based hierarchical autoencoder for nonlinear mode decomposition of fluid field data, Physics of Fluids 32~(9) (2020).

\bibitem{fukami2021model}
K.~Fukami, K.~Hasegawa, T.~Nakamura, M.~Morimoto, K.~Fukagata, Model order reduction with neural networks: Application to laminar and turbulent flows, SN Computer Science 2~(6) (2021) 467.

\bibitem{nakamura2021convolutional}
T.~Nakamura, K.~Fukami, K.~Hasegawa, Y.~Nabae, K.~Fukagata, Convolutional neural network and long short-term memory based reduced order surrogate for minimal turbulent channel flow, Physics of Fluids 33~(2) (2021).

\bibitem{sakurada2014anomaly}
M.~Sakurada, T.~Yairi, Anomaly detection using autoencoders with nonlinear dimensionality reduction, in: Proceedings of the MLSDA 2014 2nd workshop on machine learning for sensory data analysis, 2014, pp. 4--11.

\bibitem{cheng2020advanced}
M.~Cheng, F.~Fang, C.~Pain, I.~Navon, An advanced hybrid deep adversarial autoencoder for parameterized nonlinear fluid flow modelling, Computer Methods in Applied Mechanics and Engineering 372 (2020) 113375.

\bibitem{qu2021deep}
J.~Qu, W.~Cai, Y.~Zhao, Deep learning method for identifying the minimal representations and nonlinear mode decomposition of fluid flows, Physics of Fluids 33~(10) (2021).

\bibitem{cao2024laplace}
Q.~Cao, S.~Goswami, G.~E. Karniadakis, Laplace neural operator for solving differential equations, Nature Machine Intelligence 6~(6) (2024) 631--640.

\bibitem{li2020fourier}
Z.~Li, N.~Kovachki, K.~Azizzadenesheli, B.~Liu, K.~Bhattacharya, A.~Stuart, A.~Anandkumar, Fourier neural operator for parametric partial differential equations, arXiv:2010.08895 (2020).

\bibitem{raonic2023convolutional}
B.~Raonic, R.~Molinaro, T.~Rohner, S.~Mishra, E.~de~Bezenac, Convolutional neural operators, in: ICLR 2023 workshop on physics for machine learning, 2023.

\bibitem{lu2019deeponet}
L.~Lu, P.~Jin, G.~E. Karniadakis, Deeponet: Learning nonlinear operators for identifying differential equations based on the universal approximation theorem of operators, arXiv:1910.03193 (2019).

\bibitem{lu2022comprehensive}
L.~Lu, X.~Meng, S.~Cai, Z.~Mao, S.~Goswami, Z.~Zhang, G.~E. Karniadakis, A comprehensive and fair comparison of two neural operators (with practical extensions) based on fair data, Computer Methods in Applied Mechanics and Engineering 393 (2022) 114778.

\bibitem{rahman2022generative}
M.~A. Rahman, M.~A. Florez, A.~Anandkumar, Z.~E. Ross, K.~Azizzadenesheli, Generative adversarial neural operators, arXiv:2205.03017 (2022).

\bibitem{lim2023score}
J.~H. Lim, N.~B. Kovachki, R.~Baptista, C.~Beckham, K.~Azizzadenesheli, J.~Kossaifi, V.~Voleti, J.~Song, K.~Kreis, J.~Kautz, et~al., Score-based diffusion models in function space, arXiv:2302.07400 (2023).

\bibitem{seidman2023variational}
J.~H. Seidman, G.~Kissas, G.~J. Pappas, P.~Perdikaris, Variational autoencoding neural operators, arXiv:2302.10351 (2023).

\bibitem{li2020multipole}
Z.~Li, N.~Kovachki, K.~Azizzadenesheli, B.~Liu, A.~Stuart, K.~Bhattacharya, A.~Anandkumar, Multipole graph neural operator for parametric partial differential equations, Advances in Neural Information Processing Systems 33 (2020) 6755--6766.

\bibitem{li2022transformer}
Z.~Li, K.~Meidani, A.~B. Farimani, Transformer for partial differential equations' operator learning, arXiv:2205.13671 (2022).

\bibitem{serrano2023operator}
L.~Serrano, L.~Le~Boudec, A.~Kassa{\"\i}~Koupa{\"\i}, T.~X. Wang, Y.~Yin, J.-N. Vittaut, P.~Gallinari, {Operator learning with neural fields: Tackling PDEs on general geometries}, Advances in Neural Information Processing Systems 36 (2023) 70581--70611.

\bibitem{hao2023gnot}
Z.~Hao, Z.~Wang, H.~Su, C.~Ying, Y.~Dong, S.~Liu, Z.~Cheng, J.~Song, J.~Zhu, {GNOT: A general neural operator transformer for operator learning}, in: International Conference on Machine Learning, PMLR, 2023, pp. 12556--12569.

\bibitem{alkin2024universal}
B.~Alkin, A.~F{\"u}rst, S.~Schmid, L.~Gruber, M.~Holzleitner, J.~Brandstetter, Universal physics transformers: A framework for efficiently scaling neural operators, Advances in Neural Information Processing Systems 37 (2024) 25152--25194.

\bibitem{li2023geometry}
Z.~Li, N.~Kovachki, C.~Choy, B.~Li, J.~Kossaifi, S.~Otta, M.~A. Nabian, M.~Stadler, C.~Hundt, K.~Azizzadenesheli, et~al., {Geometry-informed neural operator for large-scale 3D PDEs}, Advances in Neural Information Processing Systems 36 (2023) 35836--35854.

\bibitem{han2025geomano}
X.~Han, J.~Zhang, D.~Samaras, F.~Hou, H.~Qin, {GeoMaNO: Geometric Mamba Neural Operator for Partial Differential Equations}, arXiv:2505.12020 (2025).

\bibitem{bukka2021assessment}
S.~R. Bukka, R.~Gupta, A.~R. Magee, R.~K. Jaiman, Assessment of unsteady flow predictions using hybrid deep learning based reduced-order models, Physics of Fluids 33~(1) (2021).

\bibitem{gupta2022three}
R.~Gupta, R.~Jaiman, {Three-dimensional deep learning-based reduced order model for unsteady flow dynamics with variable Reynolds number}, Physics of Fluids 34~(3) (2022).

\bibitem{chen2018neural}
R.~T. Chen, Y.~Rubanova, J.~Bettencourt, D.~K. Duvenaud, Neural ordinary differential equations, Advances in neural information processing systems 31 (2018).

\bibitem{zhang2023nonlinear}
B.~Zhang, Nonlinear mode decomposition via physics-assimilated convolutional autoencoder for unsteady flows over an airfoil, Physics of Fluids 35~(9) (2023).

\bibitem{peng2022attention}
W.~Peng, Z.~Yuan, J.~Wang, Attention-enhanced neural network models for turbulence simulation, Physics of Fluids 34~(2) (2022).

\bibitem{tripura2022wavelet}
T.~Tripura, S.~Chakraborty, Wavelet neural operator: a neural operator for parametric partial differential equations, arXiv:2205.02191 (2022).

\bibitem{rahman2022u}
M.~A. Rahman, Z.~E. Ross, K.~Azizzadenesheli, {U-NO: U-shaped neural operators}, arXiv:2204.11127 (2022).

\bibitem{chen2024learning}
G.~Chen, X.~Liu, Q.~Meng, L.~Chen, C.~Liu, Y.~Li, Learning neural operators on riemannian manifolds, National Science Open 3~(6) (2024) 20240001.

\bibitem{fu2025spatio}
X.~Fu, G.~Chen, Y.~Li, X.~Liu, L.~Chen, Q.~Meng, C.~Liu, X.~Hao, Spatio-temporal neural operator on complex geometries, Computer Physics Communications (2025) 109754.

\bibitem{peng2023linear}
W.~Peng, Z.~Yuan, Z.~Li, J.~Wang, Linear attention coupled fourier neural operator for simulation of three-dimensional turbulence, Physics of Fluids 35~(1) (2023).

\bibitem{ye2025recurrent}
Z.~Ye, C.-S. Zhang, W.~Wang, {Recurrent Neural Operators: Stable Long-Term PDE Prediction}, arXiv:2505.20721 (2025).

\bibitem{karniadakis2021physics}
G.~E. Karniadakis, I.~G. Kevrekidis, L.~Lu, P.~Perdikaris, S.~Wang, L.~Yang, Physics-informed machine learning, Nature Reviews Physics 3~(6) (2021) 422--440.

\bibitem{faroughi2022physics}
S.~A. Faroughi, N.~Pawar, C.~Fernandes, M.~Raissi, S.~Das, N.~K. Kalantari, S.~K. Mahjour, Physics-guided, physics-informed, and physics-encoded neural networks in scientific computing, arXiv:2211.07377 (2022).

\bibitem{raissi2019physics}
M.~Raissi, P.~Perdikaris, G.~E. Karniadakis, Physics-informed neural networks: A deep learning framework for solving forward and inverse problems involving nonlinear partial differential equations, Journal of Computational physics 378 (2019) 686--707.

\bibitem{mattey2021physics}
R.~Mattey, S.~Ghosh, A physics informed neural network for time-dependent nonlinear and higher order partial differential equations, arXiv:2106.07606 (2021).

\bibitem{arzani2023theory}
A.~Arzani, K.~W. Cassel, R.~M. D'Souza, Theory-guided physics-informed neural networks for boundary layer problems with singular perturbation, Journal of Computational Physics 473 (2023) 111768.

\bibitem{mohan2023embedding}
A.~T. Mohan, N.~Lubbers, M.~Chertkov, D.~Livescu, Embedding hard physical constraints in neural network coarse-graining of three-dimensional turbulence, Physical Review Fluids 8~(1) (2023) 014604.

\bibitem{chalapathi2024scaling}
N.~Chalapathi, Y.~Du, A.~Krishnapriyan, Scaling physics-informed hard constraints with mixture-of-experts, arXiv:2402.13412 (2024).

\bibitem{karnakov2024solving}
P.~Karnakov, S.~Litvinov, P.~Koumoutsakos, Solving inverse problems in physics by optimizing a discrete loss: Fast and accurate learning without neural networks, PNAS nexus 3~(1) (2024) pgae005.

\bibitem{wiewel2019latent}
S.~Wiewel, M.~Becher, N.~Thuerey, Latent space physics: Towards learning the temporal evolution of fluid flow, in: Computer graphics forum, Vol.~38, Wiley Online Library, 2019, pp. 71--82.

\bibitem{wu2022learning}
T.~Wu, T.~Maruyama, J.~Leskovec, Learning to accelerate partial differential equations via latent global evolution, Advances in Neural Information Processing Systems 35 (2022) 2240--2253.

\bibitem{brandstetter2022message}
J.~Brandstetter, D.~Worrall, M.~Welling, {Message passing neural PDE solvers}, arXiv:2202.03376 (2022).

\bibitem{lippe2023pde}
P.~Lippe, B.~Veeling, P.~Perdikaris, R.~Turner, J.~Brandstetter, {PDE-refiner: Achieving accurate long rollouts with neural PDE solvers}, Advances in Neural Information Processing Systems 36 (2023) 67398--67433.

\bibitem{liu2024multi}
X.-Y. Liu, M.~Zhu, L.~Lu, H.~Sun, J.-X. Wang, Multi-resolution partial differential equations preserved learning framework for spatiotemporal dynamics, Communications Physics 7~(1) (2024) 31.

\bibitem{li2025latent}
Z.~Li, S.~Patil, F.~Ogoke, D.~Shu, W.~Zhen, M.~Schneier, J.~R. Buchanan~Jr, A.~B. Farimani, {Latent neural PDE solver: A reduced-order modeling framework for partial differential equations}, Journal of Computational Physics 524 (2025) 113705.

\bibitem{fukami2023grasping}
K.~Fukami, K.~Taira, Grasping extreme aerodynamics on a low-dimensional manifold, Nature Communications 14~(1) (2023) 6480.

\bibitem{mousavi2025low}
H.~Mousavi, J.~D. Eldredge, Low-order flow reconstruction and uncertainty quantification in disturbed aerodynamics using sparse pressure measurements, Journal of Fluid Mechanics 1013 (2025) A41.

\bibitem{fukami2025observable}
K.~Fukami, K.~Taira, Observable-augmented manifold learning for multi-source turbulent flow data, Journal of Fluid Mechanics 1010 (2025) R4.

\bibitem{boral2023neural}
A.~Boral, Z.~Y. Wan, L.~Zepeda-N{\'u}{\~n}ez, J.~Lottes, Q.~Wang, Y.-f. Chen, J.~Anderson, F.~Sha, Neural ideal large eddy simulation: Modeling turbulence with neural stochastic differential equations, Advances in neural information processing systems 36 (2023) 69270--69283.

\bibitem{sirignano2023dynamic}
J.~Sirignano, J.~F. MacArt, Dynamic deep learning les closures: Online optimization with embedded dns, arXiv:2303.02338 (2023).

\bibitem{kochkov2021machine}
D.~Kochkov, J.~A. Smith, A.~Alieva, Q.~Wang, M.~P. Brenner, S.~Hoyer, Machine learning--accelerated computational fluid dynamics, Proceedings of the National Academy of Sciences 118~(21) (2021) e2101784118.

\bibitem{fan2024differentiable}
X.~Fan, J.-X. Wang, Differentiable hybrid neural modeling for fluid-structure interaction, Journal of Computational Physics 496 (2024) 112584.

\bibitem{tompson2017accelerating}
J.~Tompson, K.~Schlachter, P.~Sprechmann, K.~Perlin, Accelerating eulerian fluid simulation with convolutional networks, in: International Conference on Machine Learning, PMLR, 2017, pp. 3424--3433.

\bibitem{duraisamy2019turbulence}
K.~Duraisamy, G.~Iaccarino, H.~Xiao, Turbulence modeling in the age of data, Annual Review of Fluid Mechanics 51~(1) (2019) 357--377.

\bibitem{vinuesa2022enhancing}
R.~Vinuesa, S.~L. Brunton, Enhancing computational fluid dynamics with machine learning, Nature Computational Science 2~(6) (2022) 358--366.

\bibitem{margenberg2024dnn}
N.~Margenberg, R.~Jendersie, C.~Lessig, T.~Richter, {DNN-MG: A hybrid neural network/finite element method with applications to 3D simulations of the Navier--Stokes equations}, Computer Methods in Applied Mechanics and Engineering 420 (2024) 116692.

\bibitem{belbute2020combining}
F.~D.~A. Belbute-Peres, T.~Economon, Z.~Kolter, Combining differentiable pde solvers and graph neural networks for fluid flow prediction, in: International Conference on Machine Learning, PMLR, 2020, pp. 2402--2411.

\bibitem{list2022learned}
B.~List, L.-W. Chen, N.~Thuerey, Learned turbulence modelling with differentiable fluid solvers: physics-based loss functions and optimisation horizons, Journal of Fluid Mechanics 949 (2022) A25.

\bibitem{list2025differentiability}
B.~List, L.-W. Chen, K.~Bali, N.~Thuerey, Differentiability in unrolled training of neural physics simulators on transient dynamics, Computer Methods in Applied Mechanics and Engineering 433 (2025) 117441.

\bibitem{akhare2025implicit}
D.~Akhare, P.~Du, T.~Luo, J.-X. Wang, Implicit neural differential model for spatiotemporal dynamics, arXiv:2504.02260 (2025).

\bibitem{wang2024beyond}
C.~Wang, J.~Berner, Z.~Li, D.~Zhou, J.~Wang, J.~Bae, A.~Anandkumar, Beyond closure models: Learning chaotic-systems via physics-informed neural operators, arXiv:2408.05177 (2024).

\bibitem{fries2022lasdi}
W.~D. Fries, X.~He, Y.~Choi, {LaSDI: Parametric latent space dynamics identification}, Computer Methods in Applied Mechanics and Engineering 399 (2022) 115436.

\bibitem{champion2019data}
K.~Champion, B.~Lusch, J.~N. Kutz, S.~L. Brunton, Data-driven discovery of coordinates and governing equations, Proceedings of the National Academy of Sciences 116~(45) (2019) 22445--22451.

\bibitem{reinbold2019data}
P.~A. Reinbold, R.~O. Grigoriev, Data-driven discovery of partial differential equation models with latent variables, Physical Review E 100~(2) (2019) 022219.

\bibitem{maulik2020time}
R.~Maulik, A.~Mohan, B.~Lusch, S.~Madireddy, P.~Balaprakash, D.~Livescu, Time-series learning of latent-space dynamics for reduced-order model closure, Physica D: Nonlinear Phenomena 405 (2020) 132368.

\bibitem{huang2020learning}
D.~Z. Huang, K.~Xu, C.~Farhat, E.~Darve, Learning constitutive relations from indirect observations using deep neural networks, Journal of Computational Physics 416 (2020) 109491.

\bibitem{negiar2022learning}
G.~N{\'e}giar, M.~W. Mahoney, A.~S. Krishnapriyan, Learning differentiable solvers for systems with hard constraints, arXiv:2207.08675 (2022).

\bibitem{park2024tlasdi}
J.~S.~R. Park, S.~W. Cheung, Y.~Choi, Y.~Shin, {tLaSDI: Thermodynamics-informed latent space dynamics identification}, Computer Methods in Applied Mechanics and Engineering 429 (2024) 117144.

\bibitem{bonneville2024comprehensive}
C.~Bonneville, X.~He, A.~Tran, J.~S. Park, W.~Fries, D.~A. Messenger, S.~W. Cheung, Y.~Shin, D.~M. Bortz, D.~Ghosh, et~al., A comprehensive review of latent space dynamics identification algorithms for intrusive and non-intrusive reduced-order-modeling, arXiv:2403.10748 (2024).

\bibitem{dashtbayaz2025physics}
N.~H. Dashtbayaz, H.~Salehipour, A.~Butscher, N.~Morris, Physics-informed reduced order modeling of time-dependent pdes via differentiable solvers, arXiv:2505.14595 (2025).

\bibitem{paliard2022exploring}
C.~Paliard, N.~Thuerey, K.~Um, Exploring physical latent spaces for high-resolution flow restoration, arXiv:2211.11298 (2022).

\bibitem{sengupta2003analysis}
T.~K. Sengupta, G.~Ganeriwal, S.~De, Analysis of central and upwind compact schemes, Journal of Computational Physics 192~(2) (2003) 677--694.

\bibitem{logg2012automated}
A.~Logg, K.-A. Mardal, G.~Wells, Automated solution of differential equations by the finite element method: The FEniCS book, Vol.~84, Springer Science \& Business Media, 2012.

\bibitem{mahmoudi2021story}
M.~Mahmoudi, A.~Farghadan, D.~R. McConnell, A.~J. Barker, J.~J. Wentzel, M.~J. Budoff, A.~Arzani, The story of wall shear stress in coronary artery atherosclerosis: biochemical transport and mechanotransduction, Journal of biomechanical engineering 143~(4) (2021) 041002.

\bibitem{kim2010patient}
H.~J. Kim, I.~Vignon-Clementel, J.~Coogan, C.~Figueroa, K.~Jansen, C.~Taylor, Patient-specific modeling of blood flow and pressure in human coronary arteries, Annals of biomedical engineering 38~(10) (2010) 3195--3209.

\bibitem{updegrove2017simvascular}
A.~Updegrove, N.~M. Wilson, J.~Merkow, H.~Lan, A.~L. Marsden, S.~C. Shadden, Simvascular: an open source pipeline for cardiovascular simulation, Annals of biomedical engineering 45~(3) (2017) 525--541.

\bibitem{singh2004energy}
S.~P. Singh, S.~Mittal, Energy spectra of flow past a circular cylinder, International Journal of Computational Fluid Dynamics 18~(8) (2004) 671--679.

\bibitem{adam_sindy}
S.~Viknesh, Y.~Tatari, C.~Christenson, A.~Arzani, {ADAM-SINDy: An efficient optimization framework for parameterized nonlinear dynamical system identification}, Physical Review Research 8 (2026) 013040.

\bibitem{bunker2025autoencoders}
J.~Bunker, M.~Girolami, H.~Lambley, A.~M. Stuart, T.~Sullivan, Autoencoders in function space, Journal of Machine Learning Research 26~(165) (2025) 1--54.

\bibitem{csala2025physics}
H.~Csala, A.~Mohan, D.~Livescu, A.~Arzani, Physics-constrained coupled neural differential equations for one dimensional blood flow modeling, Computers in Biology and Medicine 186 (2025) 109644.

\bibitem{lehmann2025multiple}
F.~Lehmann, F.~Gatti, D.~Clouteau, Multiple-input fourier neural operator (mifno) for source-dependent 3d elastodynamics, Journal of Computational Physics 527 (2025) 113813.

\bibitem{jin2022mionet}
P.~Jin, S.~Meng, L.~Lu, Mionet: Learning multiple-input operators via tensor product, SIAM Journal on Scientific Computing 44~(6) (2022) A3490--A3514.

\bibitem{sun2024lemon}
J.~Sun, Z.~Zhang, H.~Schaeffer, Lemon: Learning to learn multi-operator networks, arXiv:2408.16168 (2024).

\bibitem{murata2020nonlinear}
T.~Murata, K.~Fukami, K.~Fukagata, Nonlinear mode decomposition with convolutional neural networks for fluid dynamics, Journal of Fluid Mechanics 882 (2020) A13.

\bibitem{morimoto2021convolutional}
M.~Morimoto, K.~Fukami, K.~Zhang, A.~G. Nair, K.~Fukagata, Convolutional neural networks for fluid flow analysis: toward effective metamodeling and low dimensionalization, Theoretical and Computational Fluid Dynamics 35~(5) (2021) 633--658.

\end{thebibliography}

\end{document}